\documentclass{clv2025}
\usepackage{hyperref}
\hypersetup{colorlinks,allcolors=black}
\usepackage{xcolor}
\usepackage{footmisc}
\usepackage{graphicx}
\usepackage{amsmath}
\usepackage{verbatim}
\usepackage{setspace}
\usepackage{multirow}
\usepackage{pifont}

\definecolor{darkblue}{rgb}{0, 0, 0.5}

\hypersetup{colorlinks=true,citecolor=darkblue, linkcolor=darkblue, urlcolor=darkblue}
\usepackage{marginnote}

\newcommand{\sg}[1]{\textcolor{blue}{#1-sg}}
\newcommand{\pl}[1]{\textcolor{red}{#1-pl}}
\newcommand{\ul}{\underline}

\makeatletter\newdimen\@footmax\makeatother
\begin{document}
\issue{1}{1}{2016}

\runningtitle{BLMs to Investigate the Linguistic Competence of LMs}

\runningauthor{Merlo, Jiang, Samo, Nastase}

\title{Blackbird Language Matrices: A Framework to Investigate the Linguistic Competence of Language Models}

\author{Paola Merlo$^{1,2}$, Chunyang Jiang$^{1,2}$, Giuseppe Samo$^{1}$, and Vivi Nastase$^{1}$}

\affilblock{
\affil{Idiap Research Institute}
\affil{University of Geneva}
}

\maketitle

\begin{abstract}

This article describes a novel language task, the  Blackbird Language Matrices (BLM) task, inspired by intelligence tests, and illustrates  the BLM datasets, their construction and benchmarking, and targeted experiments on chunking and systematicity. 
BLMs are multiple-choice problems, structured at multiple levels: within each sentence, across the input sequence, within each candidate answer. Because of their rich structure, 
these curated, but naturalistic datasets are key to answer some core questions about current large language models abilities:
do LLMs detect linguistic objects and their properties? Do they detect and use systematic patterns across sentences? Are they more prone to linguistic or reasoning errors, and how do these interact?

We show that BLMs, while challenging, can be solved at good levels of performance, in more than one language, with simple baseline models or, at better performance levels, with more tailored models. 
We show that their representations contain  the grammatical objects and attributes relevant to solve a linguistic task. We also show that these solutions are reached by detecting systematic patterns across sentences.   

The paper supports the point of view that curated, structured datasets support multi-faceted investigations of properties of language and large language models. 
Because they present a curated, articulated structure, because they comprise both learning contexts and expected answers, and because they are partly built by hand, BLMs fall in the category of datasets that can support explainability investigations, and be useful to ask \textit{why} large language models behave the way they do.
\end{abstract}


\pagebreak 

\pagebreak

\section{Introduction}
Current generative large language models translate across close languages very well, produce fluent and informative summaries, answer complex questions. These are only some of the many tasks performed with such ease and fluidity that they create a deep-seated subjective impression of human-like interaction and intelligence and an objectively useful set of tools for many practically important linguistic tasks.

And yet, large language models still struggle to handle low-resource languages, apparently relying on English as a pivot internal language \cite{wendler-etal-2024-llamas}, still do not work well on tasks for which large amounts of data are not available, are brittle, incoherent, sycophantic \cite{sharma2025sycophancy}, often factually incorrect \cite{huang2025hallucination}. 
More than anything, their non-human limitations are shown by the fact that their basic linguistic skills require computationally expensive algorithms and prohibitively large amounts of training data that are available for only a few, non-representative languages.

In short, large language models do not generalise nor abstract well or systematically. 
%
Humans, instead are good at abstraction and generalisation.
One likely reason why people generalise well is that they have a strong prior bias, grounded in the actual structure of the problem to be solved. 
A large body of literature of experimental work has demonstrated that the human mind is predisposed to extract regularities and generate rules from data, in a way that is distinct from the patterns of activation of neural networks \citep{lakretz2019emergence,lakretz2021mechanisms}.  
%
To reach better, possibly human-like, abilities in neural networks' systematic abstraction and generalisation, we need  to develop tasks and data that help us understand their current generalisation abilities and help us train them to more complex skills. 

One possible approach to develop more robust methods, then, is to pay more attention to the decomposition of complex observations 
\citep{scholkopf-etal2012}. 
Let's look at an illustrative example of complex argument structure relations in the lexicon:  the \textit{spray/load alternation} in English, shown below.


\begin{tabular}{ll|ll}
\\
The truck was loaded &  with hay.    & The hay was loaded   & on the truck.\\
    \textsc{\small  Locative}             & \textsc{\small  Instrumental}       &{\small \textsc{Theme}    }            & \textsc{\small  Locative}\\\\
\end{tabular}

This alternation applies to verbs such as {\it spray, paint, spread, fill, stuff} and {\it load}, verbs that describe covering surfaces or filling volumes.
They occur in two subcategorisation frames that are related to each other in a regular way: the object of the preposition {\it  with}  is the subject of the {\it onto} frame, while the object of the {\it onto} prepositional phrase is the subject of the {\it with} frame.
To learn the structure of such a complex alternation automatically, a neural network must be able to identify  the elements  that undergo the alternation, and their relevant attributes, and recognize the operations that manipulate these objects, across more than one sentence. 
%

%
The neural network needs to exhibit properties typical of  natural linguistic generalisation, the rules of grammar: object induction, structure-dependency and compositional systematicity. In the provided example, to learn the linguistic phenomenon, one must identify  constituents (the NPs, PPs and the verb, this is object induction). One must then identify their relations in a sentence (for example,  the mapping of their grammatical function and their semantic role, this is structure dependency); and, finally, one must identify  the systematic correspondence  of the compositional elements, the constituents, across sentences in the paradigm (for example, the crossing of grammatical functions and semantic roles across the two paradigms, this is compositional systematicity).

To investigate systematic generalisation in neural networks, we take a data-centric approach
and develop curated synthetic data on a large scale. We develop data and diagnostic probes that help us understand the network's current generalisation abilities ---what exactly do LLMs understand of the language they produce and process so well?   

Inspired by the IQ test Raven Progressive Matrices (RPMs) \cite{raven1938}, we have developed a new linguistic task, which we call Blackbird Language Matrices (BLMs). BLMs define  a prediction task to learn complex linguistic  patterns and paradigms \cite{merlo2023}. Section 2 introduces the BLM problem \cite{merlo2023} and the French example from the Agremeent problem, as in \citet{an-etal-2023-blm}. 
Sections 3 and 4 organise and systematize several  BLM templates and datasets: Agreement \cite{an-etal-2023-blm} in French and its multilingual version in \citet{nastase-etal-2024-multilingual}); the \textit{spray/load} BLM template \cite{samo-etal-2023-blm}; the Change of State and object drop verbs for Italian and English \cite{nastase-etal-2024-exploring,samo-merloLREC2026}.  Section 4.5 introduces a novel validation for the data generation process.

The novel task, and the linguistically-informed structured datasets that define it, support rich multiple levels of analysis in a single framework and allow the study of LLM representations in a systematic way. 
%
One of the distinctive characteristics of the BLMs is that one can use them to probe models at multiple levels, to get a more precise picture of what they have  actually discovered, what has emerged, and what kind of structures they assemble. 
Based on new variants of previously investigated baseline architectures and baselines \cite{samo-etal-2023-blm,nastase-merlo-2023}, and new decoder-derived sentence embeddings, we specifically investigate object induction (section 6). We use BLMs to ask what kind of linguistic objects the LLM actually learn and manipulate: Are they only tokens? Does the LLM discover something more linguistically justified like phrases and constituents? 
Based on a study of internal structure and structure dependency \cite{nastase2024identifiable} and new sentence embedding representations, in section 7 we use BLMs to investigate whether the inner representations developed by NNs represent information about chunks and if they are mapped to semantic roles, or if long-distance dependencies such as agreement can be represented. 
In a novel study of systematicity (section 8), we ask if the object and attributes are represented by higher-level abstractions or if they are based on low-level cues. Do the abstractions hold across languages?
To answer these questions, we develop BLM data  with attention to a variety of phenomena and languages.


As argued in the introduction, research questions and related work section, 
BLMs share objectives with several current learning and diagnostic datasets. 
Like ARC \cite{chollet2019}, BLMs are built to study generalisation and analogical reasoning abilities, but unlike ARC they correspond to natural puzzles, language paradigms, those language puzzles the child learning their first language has to solve or the linguist documenting the rules underlying the patterns of novel languages has to document.
BLMs share the contrastive techniques with BLiMP \cite{warstadt-etal-2020-blimp-benchmark}, a dataset of minimal sentence pairs, where each pair embodies one property of one language phenomenon. Unlike BLiMP though, the BLM puzzle is complex and cannot be solved by a simple binary classifier.
Compared to Holmes \cite{waldis-etal-2024-holmes}, BLMs can be used in a  systematic way, to investigate language properties at different levels of structure and representations, within one puzzle and across puzzles.
Compared to Digital matrices \cite{webb2023emergent}, BLMs are not simple translitterations of visual stimuli for textual inputs. They are true analogies of the visual RPMs, as they translate the concepts in RPMs into linguistic stimuli in a way that embodies the higher-level properties of language.

The larger contribution of the work lies in the definition of a new challenging learning task, in  the development  of novel structured data on different linguistic problems for several diverse languages, and in the formalisation of a general procedure to develop many other such datasets.   Most importantly,  it also lies in tackling a mixture of language tasks and reasoning that takes us closer to investigations of human linguistic intelligence.\footnote{
The datasets are available at \url{https://www.idiap.ch/en/scientific-research/data}, 
where they can be found by searching with the string `BLM'. 
The code is available at \url{https://github.com/CLCL-Geneva/BLM-SNFDisentangling}.
}

\section{The BLM task: a new structured task}

\begin{figure}
\centering
\includegraphics[width=0.6\linewidth]{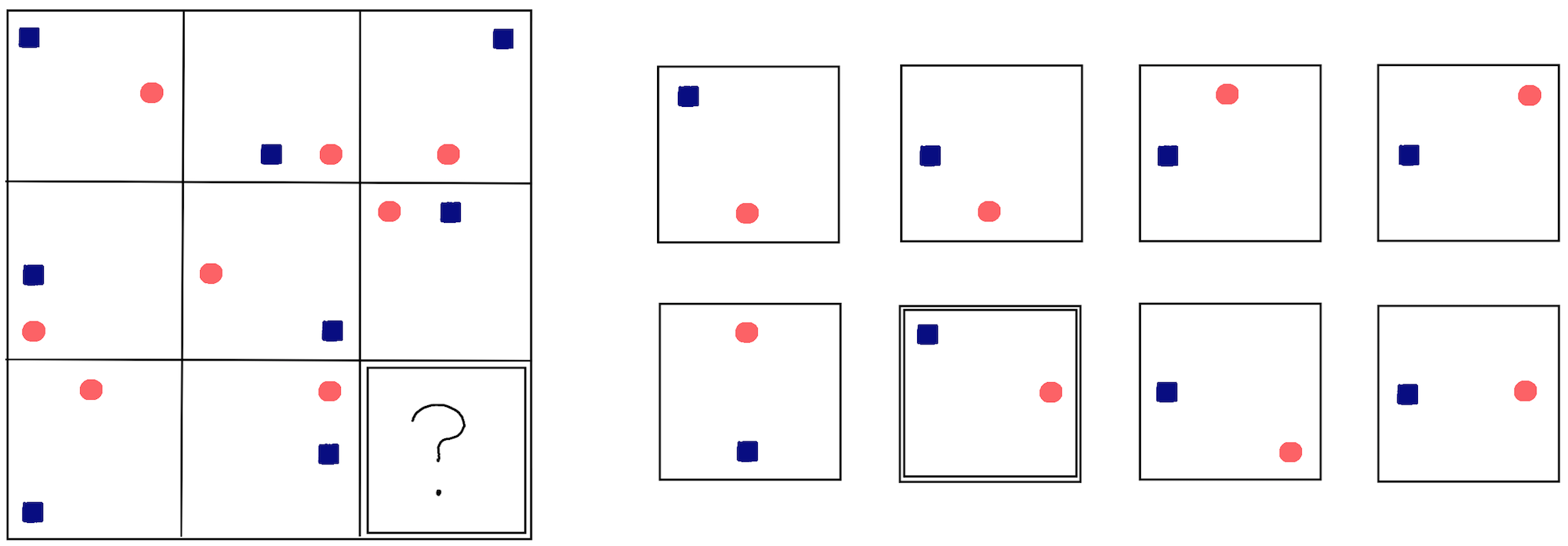}
  \caption{Example of a Raven's Progressive Matrix (RPM) from visual intelligence tests. This instance is generated with two  generative rules: (i) the red dot moves one place clockwise when traversing the matrix left to right; (ii) the blue square moves one place anticlockwise when traversing the matrix top to bottom. The task consists in finding the tile in the answer set that correctly completes the sequence (indicated with a double border).}
 \label{RPM}
\end{figure}

Blackbird Language Matrices (BLMs) are linguistic puzzles developed in analogy to the visual Raven Progressive Matrices tests \cite{merlo2023}. Raven's Progressive Matrices (RPMs) consist of a sequence of images, called the \textit{context}, connected in a logical sequence by underlying generative rules \citep{raven1938}.  
The task is to determine the  missing element in this visual sequence, the \textit{answer}, chosen among a set of closely or loosely similar alternatives, as illustrated in Figure \ref{RPM}.

Unlike other attempts to create textual versions of RPMs, BLMs are not simplistic transcriptions of visual stimuli  \cite{webb2023emergent} ---a technique that, in practice, might give away parts of the solution to the problem---, nor are  they auxiliary abstractions of stimuli in the visual domain \cite{hu-etal-2023-context}. Instead, BLMs are matrices developed specifically to learn language-related problems and delve into deeper formal and semantic properties of language, through a process of linguistic paradigm understanding.

Like RPMs, a BLM instance consists of a context set and an answer set. The context is a sequence of sentences that encode a linguistic pattern which needs to be identified in order to predict the answer, as the answer  is the continuation of this pattern.  The pattern is in terms of linguistic concepts and their rules, so it can only be identified by a system whose internal representations include the concepts and rules.  The minimally contrastive alternatives in the answer set are designed to identify specific mistakes which could be made in this process of identifying the correct concepts and patterns. More formal specifications of BLMs and their properties are defined  in Figure~\ref{fig:formal-specs}. The figure provides the formal definitions of the concepts we will use in the paper, in defining the BLM datasets, templates and problems.

\begin{figure}[h!]
    \centering
    \input{formaldefs-summary}
    \caption{Formal definitions of Blackbird Language Matrices}
    \label{fig:formal-specs}
\end{figure}

An example template is illustrated in Figure \ref{fig:template-matrices}
 \cite{an-etal-2023-blm}.
In this example, the BLM encodes  the rule of grammatical number concord: subject and verb  agree in their grammatical number, and they do so independently of how many noun phrases intervene between them.  In order to examine the representations underlying the response, and to understand what kind of linguistic information the LLMs encode, the answer sets include not only the correct answer, but also erroneous candidates constructed by corrupting the generating rules.

\begin{figure}
\footnotesize
\setlength{\tabcolsep}{2pt}
\centering
    \begin{tabular}{lllll} 
    \hline
    \multicolumn{5}{c}{\sc Context}\\
    \hline
    1 & \sg{NP}& \sg{PP1}& & \sg{VP}  \\
    2 & \pl{NP} & \sg{PP1}& & \pl{VP}  \\
    3 & \sg{NP}& \pl{PP1} & & \sg{VP}  \\
    4 & \pl{NP} & \pl{PP1} & & \pl{VP}  \\
    5 & \sg{NP}& \sg{PP1}& \sg{PP2} &\sg{VP}  \\
    6 & \pl{NP} & \sg{PP1}  & \sg{PP2} & \pl{VP}  \\
    7 & \sg{NP}   & \pl{PP1} & \sg{PP2} &  \sg{VP}  \\
    8 & ??? & & & \\ \hline
    \end{tabular}
    \hspace{1cm}
    \begin{tabular}{lllllr} \hline
    \multicolumn{6}{c}{{\sc Answers} } \\ \hline
    1 & \pl{NP} & \pl{PP1} & \sg{PP2} & \pl{VP} & \textbf{Corr} \\
    2 & \pl{NP} & \pl{PP1} & et \sg{PP2} & \pl{VP} & Coord  \\ 
    3 & \pl{NP} & \pl{PP1} &          & \pl{VP} & WNA\\
    4 & \pl{NP} & \sg{PP1} & \sg{PP1} & \pl{VP} & WN1 \\
    5 & \pl{NP} & \pl{PP1} & \pl{PP2} & \pl{VP} & WN2 \\
    6 & \pl{NP} & \pl{PP1} & \pl{PP2} & \sg{VP} & AEV \\
    7 & \pl{NP} & \sg{PP1} & \pl{PP2} & \sg{VP} & AEN1 \\
    8 & \pl{NP} & \pl{PP1} & \sg{PP2} & \sg{VP} & AEN2 \\ \hline
     \end{tabular}

\begin{tabular}{rllll} \hline
&\multicolumn{4}{c}{\sc Context} \\ 
\hline
1 & \textcolor{blue}{The computer}   & \textcolor{blue}{with the  program}  & &\textcolor{blue}{is broken.}  \\
2 & \textcolor{red}{The computers} & \textcolor{blue}{with the  program}  & &\textcolor{red}{are broken.}  \\
3 & \textcolor{blue}{The computer}   & \textcolor{red}{with the  programs} && \textcolor{blue}{is broken.}  \\
4 & \textcolor{red}{The computers} & \textcolor{red}{with the  programs} && \textcolor{red}{are broken.}  \\
5 & \textcolor{blue}{The computer}   & \textcolor{blue}{with the  program}  & \textcolor{blue}{of the experiment} & \textcolor{blue}{is broken.}  \\
6 & \textcolor{red}{The computers} & \textcolor{blue}{with the  program}  & \textcolor{blue}{of the experiment} & \textcolor{red}{are broken.}  \\
7 & \textcolor{blue}{The computer}   & \textcolor{red}{with the  programs} & \textcolor{blue}{of the experiment} & \textcolor{blue}{is broken.}  \\
8 &  ???\\ \hline
\end{tabular}

\begin{tabular}{rl} \hline
 \multicolumn{2}{c}{\sc Answers} \\ \hline
 1 & \textcolor{red}{The computers}  \textcolor{red}{with the  programs}  \textcolor{blue}{of the experiment}  \textcolor{red}{are broken.} \\
 2 & The computers with the  programs and the experiment are broken. \\
 3 & The computer with the  programs are broken. \\
 4 & The computer with the  program of the experiment are broken. \\
 5 & The computers with the  programs of the experiments are broken. \\
 6 & The computers with the  programs of the experiments is broken. \\
 7 & The computers with the  program of the experiments is broken.\\
 8 & The computers with the  programs of the experiment is broken.\\\hline
 \end{tabular}
\caption{BLM template and instances for verb-subject agreement, with one or two intervening phrases. Three generative rules: (i) Subject matches in number with verb (singular or plural); (ii) material can intervene and is of unbounded length; (iii) singular and plurals alternate in regular patterns. The errors can be grouped in two types: (i) \textit{sequence errors}: WNA= wrong number of attractors; WN1= wrong grammmatical number for 1$^{st}$ attractor noun (N1); WN2= wrong grammatical number for 2$^{nd}$ attractor noun (N2); (ii) \textit{grammatical errors}: AEV=agreement error on the verb; AEN1=agreement error on N1; AEN2=agreement error on N2.
}
\label{fig:template-matrices}
 \end{figure}

 By their construction, BLM datasets are richly structured. A BLM instance is a multiple-choice problem 
with structure at multiple levels: nwithin each sentence,
across the input sequence,  within each candidate answer. 

Because of this multiple-level structure, BLMs support many different types of investigations, at both the sentence and matrix levels. The context-answer set up supports counterfactual investigations of possible types of errors: language errors, reasoning errors, and their interactions. As we will see in the next sections, the regular syntactic forms and the systematic semantic properties support investigations on systematicity and compositionality in neural networks (section 8). The predictable syntactic structure of individual sentences, and the structure within the sequence of a BLM context, also support investigations on how linguistic information is encoded in sentence embeddings (section 7) \cite{nastase-merlo-2024-identifiable,nastase-merlo-2024-tracking}. BLMs exist for several tasks and different languages, enabling multi-tasks and multi-language comparative studies \cite{nastase-etal-2024-multitask,nastase-etal-2024-multilingual}. Finally, each BLM problem is a linguistic paradigm and can be seen as a tool for linguistic investigation of specific  linguistic phenomena.

\section{The BLM Problems and Templates}

The creation of the different datasets follows a similar workflow, adapted to the resources available for the language under consideration.
First, the relevant linguistic phenomenon is identified. Second, the BLM problem is defined: what are the linguistic properties and the attributes that need to be learnt? Third comes the definition of the BLM template: the context and answer sets that extensionally describes the BLM problems and its properties. Then, source seed sentences need to be formulated, either by sourcing existing stimuli and examples or by creating them by  hand; they are validated by hand. Finally, a semi-automatic step augments the seed data into larger combinatorics.

\begin{table}[h]
\small
    \centering
    \begin{tabular}{l|ccc|ccc|cc|cccccc}
    &\multicolumn{3}{c}{Construction}
      &\multicolumn{3}{c}{Lexical Type}
        &\multicolumn{2}{c}{BLM Type}
          &\multicolumn{4}{c}{Languages}\\
&M & R & C & I & II & III& 1 &  2+ & E & F & I & R &T &H\\\hline      
 Caus-CoS&
\checkmark&&&\checkmark&\checkmark&\checkmark&\checkmark&&\checkmark&&\checkmark&&\checkmark&\checkmark\\
 Caus-Roll&
\checkmark&&&\checkmark&\checkmark&\checkmark&&\checkmark&\checkmark&&&&&\\
Object Drop &
\checkmark&&&\checkmark&\checkmark&\checkmark&\checkmark&\checkmark&\checkmark&&\checkmark&&&\\
Spray/load        & 
\checkmark&&&\checkmark&\checkmark&\checkmark&\checkmark&&\checkmark&&&&&\\\hline
Agr     &  \checkmark&\checkmark&\checkmark&\checkmark&\checkmark&\checkmark&\checkmark&&\checkmark&\checkmark&\checkmark&\checkmark&&\\
Seq. of Tense &
\checkmark&&& \checkmark&&&\checkmark&&&&\checkmark&&&\\
Mixture &
\checkmark&&& \checkmark&&&\checkmark&&&&\checkmark&&&\\
\end{tabular}
    \caption{Overview of properties of datasets. The datasets cover seven different linguistic paradigms: subject-verb agreement (Agr), causative alternation exemplified by change of state verbs (Caus-CoS) and \textit{Roll}-verbs (Caus-Roll), object-drop alternation, \textit{spray/load} alternation, sequence of tense and a mixture of sequence of tense and agreement. The datasets covering alternations can be used together to study contrastive properties (for example, CoS and Roll are causatives, but OD are not). The datasets are developed mostly in main sentences (M) but also in completives (C) and relatives (R). They all are available in three types, of different levels of increasingly complex lexicalisation (I, II, III). Some of the BLMs show a single eight-sentence template (1)  while others have a twice-repeated four-tuple structure (2) or even four-tuple structures. We created BLMs in several languages, with the same problems in one, two or four languages (\ul{E}nglish, \ul{F}rench, \ul{I}talian, \ul{R}omanian, \ul{T}urkish and \ul{H}ebrew), in parallel datasets. }
    \label{tab:overview}
\end{table}

\subsection{The linguistic phenomena: properties and motivations}

The linguistic phenomena for which we have built BLMs cluster around two types of properties: verb alternation phenomena and structure-dependent, long-distance phenomena. The first type is exemplified by the \textit{spray/load} alternation, the causative/inchoative alternation (Change of State verbs (CoS) and \textit{Roll }verbs), the object-drop alternation; the latter type is instantiated  by  subject-verb number agreement,  the sequence of tenses and a mixtures of the agreement and the sequence of tense problems. 

These phenomena have been chosen because they cover many aspects of language structure and knowledge. They span both the morpho-syntactic level (Agreement) and the lexical semantic interface of argument structure (Causatives, Object-drop, \textit{Spray/load}). 
Agreement and the sequence of tense problems are purely formal problems, that apply to all verbs and nouns and create a formal linear dependency between elements in a sentence. 
The problems related to verb alternations (Causatives, Object-drop, \textit{Spray/load}) instead tap into mapping lexical semantics to syntactic argument realisation and the building blocks of argument structure. The causative alternation is represented by CoS and \textit{Roll} verbs (manner of motion). They are contrasted to the Object-drop alternation.
The causative alternation and the object-drop alternation are a minimal pair for the study of the mapping of semantics to syntax and for the internal structuring of the semantics of the  lexicon.
Also, some phenomena, such as agreement or sequence of tense, are defined over a single sentence and the structure of the BLM matrix is defined as pattern of  single sentences. Others, like the verb alternations, span, by definition, a pair of sentences and can only be properly defined linguistically if the pair is identified together.
Finally, some of these problems are language-specific and are not observed productively in many languages (\textit{spray-load}), others have been found in all documented languages in some form (Causatives), others exists in many languages of a given language family (agreement), thus supporting multi-lingual comparisons.
A summary of all these different dimensions of organisation and possible comparison is given in Table \ref{tab:overview}.

\subsection{The templates}
In describing  BLMs, we  talk of \textit{contexts}, the sequence of sentences whose missing element needs to be identified, and \textit{answer set}, the  set of answers that instantiates the multiple choice task to be solved. Following the specification of Figure \ref{fig:formal-specs}, a summary of the logical structure of the BLM templates is shown in Figure  \ref{fig:logical-structure-BLM}.

\subsubsection{The \textit{spray-load} BLM template}

The \textit{spray/load} alternation is the expression of complex syntactic and semantic mapping in a verb argument structure. 
The syntactic behaviour involves an alternation in which a selected class of verbs combines three arguments to describe an event where an \textsc{Agent} causes motion of a \textsc{Theme} to a \textsc{Loc(ation)}. Both \textsc{Theme} and \textsc{Loc} can function as the syntactic direct objects of the structure, resulting in two possible alternating configurations. In one alternate, \textsc{Theme} immediately follows the verb, and \textsc{Loc} is introduced by a preposition (\textit{The student sprays the paint onto the wall}). In the other alternate, \textsc{Loc} follows the verb, and \textsc{Theme} is introduced by a preposition (\textit{The student sprays the wall with the paint}).

The matrix then is structured to show the syntactic properties of the arguments \textsc{Theme} and \textsc{Loc}, implicitly showing that the two alternates share common properties.  A  visible surface indicator of these properties is the use of the passive.
In the \textit{spray/load} verb class, the \textsc{Theme} and \textsc{Loc} can be syntactically transformed into the subject of a passive structure (e.g., \textit{paint was sprayed onto the wall} vs. \textit{the wall was sprayed with paint}, as they both share direct object properties \citep[156-159]{delia16}. 

In the answer set, the target sentence is to be chosen from a minimally contrastive set of candidates. The context is generated according to a set of rules (see Figure \ref{fig:logical-structure-BLM}) that are violated in  incorrect answers, alone or in combination with other violations: violations of the correct syntactic form of the sentence (rules 1-3);  violations of the proper lexical selection (rule 4); violations of 
proper syntax-semantic mapping (rule 5). We create two templates, one for each  of the two alternates, as shown in Figure \ref{ALT-ATL} and Figure \ref{ATL-ALT}.
Examples are shown in Figure \ref{BLMslE-typeiii}.

\begin{figure}
   \centering
   \footnotesize
\begin{tabular}{lp{0.8\linewidth}} \hline
\multicolumn{2}{c}{\textbf{Spray/load} }\\
\textbf{LP Components} & Objects= phrases (NP, PPs); \\
& Features = semantic roles (Agent, Theme, Loc); position (subject, compl).\\ \\
\textbf{Rules}
& (E)   If voice is active, agent is an NP in subject position (\textsc{AgentAct});\\ 
&(E) When PPs, arguments are introduced by given prepositions 
(\textsc{LexPrep});\\ 
&(I) The PP following the NP is not embedded in it (\textsc{NoEmb});\\ 
&(I) The order of the constituents and their role is fixed 
(\textsc{SSM});\\
& (R) Alternation between a  verb followed by an NP and a PP   (\textsc{Alt}).\\ \\
\textbf{Errors}& Violations of the correct syntactic form (rules \textsc{AgentAct,\textsc{Alt}}, \textsc{NoEmb});\\ 
&Violations of the proper lexical selection (rule \textsc{LexPrep});\\
& Violations of proper syntax-semantic mapping (rule \textsc{SSM}).\\
\hline
\multicolumn{2}{c}{\textbf{Cos/OD} }\\
\textbf{LP Components} & Objects= phrases (NP, PPs); \\
& Features = semantic roles (Agent, Theme); position (subject, object).\\\\
\textbf{Rules} & (E) the presence of one or two arguments \\
& (E) The active (\textcolor{brown}{ActiveV}) and passive (\textcolor{teal}{PassiveV}) or passive voice of the verb;\\
& (I) Argument attributes (agents, \textcolor{violet}{Agent}; patients, \textcolor{orange}{Theme});\\
& (R) Alternation between a PP introduced by any preposition and a PP introduced by a non-agentive preposition \textit{by}.\\\\
\textbf{Errors }& Violations of syntax-semantics mapping (\textsc{I-Int}, SSM).\\
& Violation of voice (\textsc{ER-Pass}).\\
& Violations of proper rule of progression (\textsc{R-Trans, ER-Pass}).\\
& Wrong lexical choice in preposition (\textsc{E-WrBy}). \\ \hline
\multicolumn{2}{c}{\textbf{Roll} }\\
\textbf{LP Components} & Objects= phrases (NP, PPs); \\
& Features = semantic roles (Agent, Theme); position (subject, object).\\\\
\textbf{Rules} & (E) The presence of one or two arguments; \\
& (I) Argument attributes (agents, \textcolor{violet}{Agent}; patients, \textcolor{orange}{Theme});\\
& (R) Repetition of quadruple (paradigms). \\\\
\textbf{Errors }& Violation of attribute mapping (Role swap, (RS), role errors (RR)).\\
& Violation of corrects paradigm (paradigm shifts, PC-RR, PSC-RR,PSC-RS).\\
& Structure changes (SC-RR,SCRS).\\  \hline
\multicolumn{2}{c}{\textbf{Agr} }\\
\textbf{LP Components} & Objects= phrases (NP, PPs); \\
& Features = grammatical number (singular, plural); feature matching (yes, no).\\\\
\textbf{Rules} & (E) Subject matches in number with verb (singular or plural);\\
&(E) Material can intervene and is of unbounded length;\\
&(I) Subject and verb are adjacent in a grammatical tree;\\
& (R) Singular and plurals alternate in regular patterns.\\ \\
\textbf{Errors}& Grammatical agreement errors (AEV=on the verb, AEN1/AEN2= on N1/N2).\\ 
& Sequence errors: Violation of the progression (WNA, WN1, WN2).\\\\\hline
\end{tabular}
    \caption{Summary of the logical structure of the BLM templates.  A BLM matrix  comprises a linguistic phenomenon LP, its objects O and their features F, external rules E, internal rules I and a progression operator R. Rules are violated in the contrastive examples of the answer set A. Multiple rules can be corrupted at the same time. 
    }
    \label{fig:logical-structure-BLM}
\end{figure}

\begin{figure}
\footnotesize
\setlength{\tabcolsep}{3pt} 
\begin{minipage}{0.49\textwidth}
\begin{tabular}{lllll} 
\hline
\multicolumn{5}{c}{\sc Context }\\
\hline
1 & \textcolor{teal}{NP-\textcolor{violet}{Agent}} & Verb & \textcolor{teal}{NP-}\textcolor{red}{Loc} & \textcolor{orange}{PP-}\textcolor{blue}{Theme} \\
2 & \textcolor{teal}{NP-\textcolor{blue}{Theme}} & VerbPass & \textcolor{orange}{PP-}\textcolor{violet}{Agent} \\
3 & \textcolor{teal}{NP-\textcolor{blue}{Theme}} & VerbPass & \textcolor{orange}{PP-}\textcolor{red}{Loc} & \textcolor{orange}{PP-}\textcolor{violet}{Agent}\\
4 & \textcolor{teal}{NP-\textcolor{blue}{Theme}} & VerbPass & \textcolor{orange}{PP-}\textcolor{red}{Loc}  \\
5 & \textcolor{teal}{NP-\textcolor{red}{Loc}} & VerbPass & \textcolor{orange}{PP-}\textcolor{violet}{Agent} \\
6 & \textcolor{teal}{NP-\textcolor{red}{Loc}} & VerbPass & \textcolor{orange}{PP-}\textcolor{blue}{Theme} & \textcolor{orange}{PP-}\textcolor{violet}{Agent} \\
7 & \textcolor{teal}{NP-\textcolor{red}{Loc}} & VerbPass & \textcolor{orange}{PP-}\textcolor{blue}{Theme} \\
8 & ??? \\\\
\end{tabular}
\end{minipage}
\begin{minipage}{0.49\textwidth}
    \vspace{-3mm}
\begin{tabular}{rll} \hline
\multicolumn{3}{c}{\sc Answers}\\ \hline
1& \textcolor{teal}{NP-\textcolor{violet}{Agent}} Verb  \textcolor{teal}{NP-}\textcolor{blue}{Theme}  \textcolor{orange}{PP-}\textcolor{red}{Loc}& \textsc{Correct} \\ 
2& NP-Agent *VerbPass NP-Theme PP-Loc & \textsc{AgentAct}\\
3& NP-Agent Verb NP-Theme *NP-Loc & \textsc{Alt-NP}\\
4& NP-Agent Verb *PP-Theme PP-Loc & \textsc{Alt-PP}\\
5& NP-Agent Verb *[NP-Theme PP-Loc] & \textsc{NoEmb}\\
6& NP-Agent Verb NP-Theme *PP-Loc & \textsc{LexPrep}\\
7& *NP-Theme Verb NP-Agent PP-Loc & \textsc{SSM-1}\\
8& *NP-Loc Verb NP-Agent PP-Theme &\textsc{SSM-2}\\
9& *NP-Theme Verb NP-Loc PP-Agent & \textsc{AASSM}
 \end{tabular}
\end{minipage}
\caption{BLM context and answers for the \textit{spray/load} alternation from \textsc{Agent}-\textsc{Loc}-\textsc{Theme} to \textsc{Agent}-\textsc{Theme}-\textsc{Loc}. \textsc{Cr} = Corrupted rule(s); * = corrupted element; brackets = syntactic embedding.
}
\label{ALT-ATL}
\end{figure}

\begin{figure}[!h]
\footnotesize
\setlength{\tabcolsep}{3pt} 
\begin{minipage}{0.49\textwidth}
    \begin{tabular}{lllll} 
    \hline
    \multicolumn{5}{c}{\sc Context}\\
    \hline
    1 & \textcolor{teal}{NP-\textcolor{violet}{Agent}} & Verb & \textcolor{teal}{NP-}\textcolor{blue}{Theme} & \textcolor{orange}{PP-}\textcolor{red}{Loc}  \\
    2 & \textcolor{teal}{NP-\textcolor{blue}{Theme}} & VerbPass & \textcolor{orange}{PP-}\textcolor{violet}{Agent} \\
    3 & \textcolor{teal}{NP-\textcolor{blue}{Theme}} & VerbPass & \textcolor{orange}{PP-}\textcolor{red}{Loc} & \textcolor{orange}{PP-}\textcolor{violet}{Agent}\\
    4 & \textcolor{teal}{NP-\textcolor{blue}{Theme}} & VerbPass & \textcolor{orange}{PP-}\textcolor{red}{Loc}  \\
    5 & \textcolor{teal}{NP-\textcolor{red}{Loc}} & VerbPass & \textcolor{orange}{PP-}\textcolor{violet}{Agent} \\
    6 & \textcolor{teal}{NP-\textcolor{red}{Loc}} & VerbPass & \textcolor{orange}{PP-}\textcolor{blue}{Theme} & \textcolor{orange}{PP-}\textcolor{violet}{Agent} \\
    7 & \textcolor{teal}{NP-\textcolor{red}{Loc}} & VerbPass & \textcolor{orange}{PP-}\textcolor{blue}{Theme} \\
    8 & ???\\\\
    \end{tabular}
\end{minipage}
\begin{minipage}{0.49\textwidth}
    \vspace{-3mm}
    \begin{tabular}{rll} \hline
    \multicolumn{3}{c}{\sc Answers}  \\ \hline
     1 & \textcolor{teal}{NP-\textcolor{violet}{Agent}}  Verb  \textcolor{teal}{NP-}\textcolor{red}{Loc}  \textcolor{orange}{PP-}\textcolor{blue}{Theme} & \textsc{Correct}  \\ 
     2 & NP-Agent *VerbPass NP-Loc PP-Theme & \textsc{AgentAct}\\
     3 & NP-Agent Verb NP-Loc *NP-Theme & \textsc{Alt-NP} \\
     4 & NP-Agent Verb *PP-Loc PP-Theme & \textsc{Alt-PP} \\
     5 & NP-Agent Verb *[NP-Loc PP-Theme] & \textsc{NoEmb} \\
     6 & NP-Agent Verb NP-Loc *PP-Theme & \textsc{LexPrep} \\
     7 & *NP-Loc Verb NP-Agent PP-Theme & \textsc{SSM-1}\\
     8 & *NP-Theme Verb NP-Agent PP-Loc & \textsc{SSM-2}\\
    9 & *NP-Loc Verb NP-Theme PP-Agent & \textsc{AASSM}\\
     \end{tabular}
\end{minipage}
\caption{BLM context and answers for the spray/load alternation from \textsc{Agent}-\textsc{Theme}-\textsc{Loc} to \textsc{Agent}-\textsc{Loc}-\textsc{Theme}.  \textsc{Cr} = Corrupted rule(s), * = locus of the rule corruption, angled brackets = syntactic embedding.
}
\label{ATL-ALT}
\end{figure}

\begin{figure}
\small
\setlength{\tabcolsep}{1mm}
\begin{tabular}{rp{0.4\linewidth}} \hline
&    \textsc{Context}\\ \hline 
1&    	\textcolor{violet}{I} pumped \textcolor{red}{the pipes} \textcolor{blue}{with regular air} \\
2&      \textcolor{blue}{The grain} was sown \textcolor{violet}{by the monks} \\
3&      \textcolor{blue}{Regular air} was pumped \textcolor{red}{into the stove} \textcolor{violet}{by me} \\
4&      \textcolor{blue}{The contents} were stuck \textcolor{red}{in a box}\\
5&      \textcolor{red}{The suitcase} was loaded \textcolor{violet}{by the buyer} \\
6&      \textcolor{red}{The dragon world} was scattered \textcolor{blue}{with monsters} \textcolor{violet}{by the wizard}\\
7&      \textcolor{red}{The surface}was splattered \textcolor{blue}{with the stuffing}\\
8&      ???\\\\ \hline    
\end{tabular}
\begin{tabular}{rp{0.5\linewidth}}\hline
 &        \sc Answers\\ \hline 
1 &    \textcolor{violet}{The tomatoes} splatter \textcolor{blue}{water} \textcolor{red}{all over the grill}\\
2 &     The scientists were seeded sulfur into the earth \\
3 &     Farmers have to squirt a tiny bit of oil everything  \\
4 &     My child smeared with the flour all over the room \\
5 &     People baste different types of marinades of seafood \\
6 &     I brushed mascara under the surface\\
7 &     The yummy almond butter spread someone all over the pastry \\
8 &     The stuff sticks Bob in the truck\\
9 &     The cart can load the supplies in the buyer\\    \hline 
\end{tabular}
\caption{Example of context sentences and answer set for the \textit{spray/load} alternation BLM.}
\label{BLMslE-typeiii}
\end{figure}

\subsubsection{The causative BLM template}

The causative alternation is represented by Change-of-State (CoS) \textit{(The teacher opened the door/The door opened)} and \textit{Roll} verbs (Manner of Motion) \textit{(The man rolled the dice/ The dice rolled)}. 

\paragraph{Change-of-State (CoS) alternation}

This alternation applies to change of state verbs, such as {\it open, break, melt, burn} among many others, verbs that describe a change that affects the state of the undergoing participant (\textit{the door} changes from a state of being closed to a state of being open).
They occur in two subcategorisation frames that are related to each other in a regular way: syntactically, the object of the transitive frame is the subject of the intransitive frame. Semantically, the transitive has a causative meaning. In terms of semantic roles, the subject of the transitive is an \textsc{agent}, but the subject of the intransitive is a \textsc{theme.}

These properties  lead to the construction of  the \textit{context set.}
Specifically, (i) the presence of one or two arguments and their attributes (agents, \textcolor{violet}{Agent}; patients, \textcolor{orange}{Theme}) ; (ii) the active (\textcolor{brown}{ActiveV}) and passive (\textcolor{teal}{PassiveV}) or passive voice of the verb. 
The factors related to the progression rule that builds the BLM involve the number and quality of nominal phrases (NP) following the verb and its core arguments. This includes an alternation between an NP introduced (i) by any preposition (e.g., \textit{in an instant}, henceforth \textcolor{blue}{p}-NP) and (ii) by the preposition by (e.g., \textit{by chance}, \textcolor{red}{by}-NP), but not agentive (e.g., \textit{by the artist}, \textcolor{red}{by}-\textcolor{violet}{NP-Agent}/\textcolor{red}{by}-\textcolor{orange}{NP-Theme}).

In the answer set, all the answers have the same structure NP V (NP) PP, consisting of a verb, two nominal constituents and a preposition (\textcolor{red}{by}, or the lack of the preposition) between the verb and the second NP, for a total of four linguistic elements (NP V \textcolor{red}{by} NP). 
The error labeled as \textsc{I-Int} represents an intransitive structure (Int). This is an error of syntax-semantics mapping, because its role is Agent. In \textsc{ER-Pass}, the verb is not in the correct voice, and is in a passive structure. \textsc{R-Trans} are violations of the operators of the BLM, and they result in a transitive sentence without a PP. Finally, \textsc{E-WrBy} are cases of ungrammatical sentences which only miminally differ from the target answer mainly by the NP which follows the preposition \textit{by}. All these errors also contain mistakes of syntax-semantic mapping.\footnote{The templates are developed for Italian and English in a parallel way. However, Italian templates have two additional answers, which pertain to the presence or absence of the reflexive \textit{si} in causative constructions.}
The templates for the context and the answer set for both sets of datasets are provided in Figure \ref{tab:COStemplateBLM} and the example sentences are in Figure \ref{tab:COSexample}.

\begin{figure}[!h]
\centering
\footnotesize 
\setlength{\tabcolsep}{2pt} 

\begin{minipage}{0.49\textwidth}
    \vspace{-3mm}
    \begin{tabular}{lllll} 
    \hline
    \multicolumn{5}{c}{\sc CoS context}\\
    \hline
    1 & \textcolor{violet}{NP-Agent} & \textcolor{brown}{ActiveV}  & \textcolor{orange}{NP-Theme} & \textcolor{blue}{P}-NP \\
    2 & \textcolor{violet}{NP-Agent} & \textcolor{brown}{ActiveV}  & \textcolor{orange}{NP-Theme}  & \textcolor{red}{by}-NP  \\
    3 & \textcolor{orange}{NP-Theme} & \textcolor{teal}{PassiveV} & \textcolor{red}{by}-\textcolor{violet}{NP-Agent} & \textcolor{blue}{P}-NP \\
    4 & \textcolor{orange}{NP-Theme} & \textcolor{teal}{PassiveV} & \textcolor{red}{by}-\textcolor{violet}{NP-Agent} & \textcolor{red}{by}-NP \\
    5 & \textcolor{orange}{NP-Theme} & \textcolor{teal}{PassiveV} & & \textcolor{blue}{P}-NP \\
    6 & \textcolor{orange}{NP-Theme} & \textcolor{teal}{PassiveV} & & \textcolor{red}{by}-NP \\
    7 & \textcolor{orange}{NP-Theme} & \textcolor{brown}{ActiveV}  & & \textcolor{blue}{P}-NP \\ 
    8 & ??? & & &  \\ 
    \end{tabular}
\end{minipage}
\hfill
\begin{minipage}{0.49\textwidth}
    \begin{tabular}{lll} \hline
    \multicolumn{3}{c}{\sc CoS answers}  \\ \hline
    1 & \textcolor{orange}{NP-Theme} \textcolor{brown}{ActiveV}   \textcolor{red}{by}-NP & \textsc{Correct}\\ 
    2 & \textcolor{violet}{NP-Agent} \textcolor{brown}{ActiveV}   \textcolor{red}{by}-NP & \textsc{I-Int}  \\ 
    3 & \textcolor{orange}{NP-Theme} \textcolor{teal}{PassiveV}  \textcolor{red}{by}-\textcolor{violet}{NP-Agent} & \textsc{ER-Pass}\\
    4 & \textcolor{violet}{NP-Agent} \textcolor{teal}{PassiveV}   \textcolor{red}{by}-\textcolor{orange}{NP-Theme} & \textsc{IER-Pass}\\
    5 & \textcolor{orange}{NP-Theme} \textcolor{brown}{ActiveV}   \textcolor{violet}{NP-Agent} & \textsc{R-Trans}\\
    6 & \textcolor{violet}{NP-Agent} \textcolor{brown}{ActiveV}   \textcolor{orange}{NP-Theme} & \textsc{IR-Trans} \\
    7 & \textcolor{orange}{NP-Theme} \textcolor{brown}{ActiveV}   \textcolor{red}{by}-\textcolor{violet}{NP-Agent} & \textsc{E-WrBy} \\
    8 & \textcolor{violet}{NP-Agent} \textcolor{brown}{ActiveV}   \textcolor{red}{by}-\textcolor{orange}{NP-Theme} & \textsc{IE-WrBy}  \\ \\ 
    \end{tabular}
\end{minipage}
\caption{ BLM template for CoS contexts and answers.
}
\label{tab:COStemplateBLM}
\end{figure}

\begin{figure}[!h]
\footnotesize
\setlength{\tabcolsep}{2pt} 
\centering
\begin{tabular}{lp{0.45\textwidth}} 
\hline
\multicolumn{2}{c}{\sc EnCoS - Context } \\
\hline
1 & \textcolor{violet}{The witch} \textcolor{brown}{breaks} \textcolor{orange}{an oath} \textcolor{blue}{within} seconds \\
2 & \textcolor{violet}{The witch}  \textcolor{brown}{breaks} \textcolor{orange}{an oath} \textcolor{red}{by} chance \\
3 & \textcolor{orange}{An oath} \textcolor{teal}{is broken} \textcolor{red}{by} \textcolor{violet}{the witch} \textcolor{blue}{within} seconds \\
4 & \textcolor{orange}{An oath} \textcolor{teal}{is broken} \textcolor{red}{by} \textcolor{violet}{the witch}  \textcolor{red}{by} chance \\
5 & \textcolor{orange}{An oath} \textcolor{teal}{is broken}  \textcolor{blue}{within} seconds \\
6 & \textcolor{orange}{An oath} \textcolor{teal}{is broken} \textcolor{red}{by} chance \\
7 & \textcolor{orange}{An oath} \textcolor{brown}{breaks} \textcolor{blue}{within} seconds\\
8 & ??? \\ 
\hline 
\end{tabular} 
\begin{tabular}{lp{0.45\textwidth}} 
\hline
\multicolumn{2}{c}{\sc EnCoS - Answers} \\
\hline
1 & \textcolor{orange}{An oath} \textcolor{brown}{breaks} \textcolor{red}{by} chance \\
2 & The witch breaks by chance \\
3 & An oath is broken by the witch \\
4 & The witch is broken by an oath \\
5 & An oath breaks the witch \\
6 & The witch breaks an oath \\
7 & An oath breaks by the witch \\
8 & The witch breaks by an oath \\
\hline
\end{tabular}
\caption{Examples of the English verb \textit{break}, one of the verbs belonging to the CoS class.}
\label{tab:COSexample}
\end{figure}

\paragraph{Roll verbs}

The \textit{Roll} verbs class comprises manner of motions verbs whose subject can be inanimate, but can have non-volitional movement (e.g \textit{roll}, \textit{wind}, etc.).
The \textit{roll} BLM template has a different structure from the previous ones. 
The \textit{roll} verb alternation template 
is composed of two quadruples, called paradigms (Figure~\ref{fig:blm-roll-eg}).
Embedded in these paradigms are soft annotations of implicit linguistic features without overt tagging. For example, the agentivity of a noun phrase (NP) might be implicitly indicated with a simple action descriptor like \textit{NP did it}, tagging it as the \textsc{Agent}. 
The associated answer set includes both correct and deliberately incorrect choices, such as role swap (RS), or more complex rule-level changes, like paradigm shifts (PS)  and structure changes (SC). 
Each incorrect answer violates exactly one constraint dimension while preserving others, enabling precise evaluation of rule component acquisition. Role errors (RR) test semantic understanding, structural errors (SC-RR, SCRS) test syntactic knowledge, and paradigm errors (PC-RR, PSC-RR, PSC-RS) test analogical consistency across contexts.

\begin{figure}
\centering
\footnotesize 
\setlength{\tabcolsep}{2pt} 

\begin{tabular}{llll} 
\hline
\multicolumn{4}{c}{\sc Context}\\
\hline
1 & \textcolor{violet}{NP-Agent} & \textcolor{blue}{rollVerb}  & \textcolor{orange}{NP-Theme} \\
2 & \textcolor{violet}{NP-Agent} & \textcolor{black}{did}  & \textcolor{orange}{NP-Theme}\\
3 & \textcolor{orange}{NP-Theme} & \textcolor{black}{was} & \textcolor{red}{NP-Loc}\\
4 & \textcolor{orange}{NP-Theme} & \textcolor{blue}{rollVerb} & \textcolor{red}{NP-Loc}\\
5 & \textcolor{violet}{NP-Agent} & \textcolor{blue}{rollVerb} & \textcolor{orange}{NP-Theme}\\
6 & \textcolor{violet}{NP-Agent} & \textcolor{black}{did} &\textcolor{orange}{NP-Theme}\\
7 & \textcolor{orange}{NP-Theme} & \textcolor{black}{was}  & \textcolor{red}{NP-Loc}\\ 
8 & ??? & &   
\end{tabular}
\hspace{0.5cm} 
\begin{tabular}{lllll} \hline
\multicolumn{3}{c}{\sc Answers}  \\ \hline
1 & \textcolor{orange}{NP-Theme} &\textcolor{blue}{rollV} &\textcolor{violet}{NP-Agent}&\textsc{Scrc}\\ 
2 & \textcolor{violet}{NP-Agent} &\textcolor{black}{was}  & \textcolor{red}{in}-\textcolor{red}{NP-Loc} & \textsc{Sc-rr} \\ 
3 & \textcolor{violet}{NP-Agent} &\textcolor{blue}{rollV}  & \textcolor{red}{in}-\textcolor{red}{NP-Loc}& \textsc{Rr }\\
4 & \textcolor{orange}{NP-Theme}& \textcolor{blue}{rollV}  & \textcolor{red}{into}-\textcolor{red}{NP-Loc} & \textsc{Correct}\\
5 & \textcolor{orange}{NP-Theme} &\textcolor{blue}{rollV}  & \textcolor{violet}{NP-Agent} & \textsc{Psc-rs}\\
6 & \textcolor{violet}{NP-Agent} &\textcolor{black}{was}   &\textcolor{red}{in}-\textcolor{red}{NP-Loc} & \textsc{Psc-rr}\\
7 & \textcolor{violet}{NP-Agent} &\textcolor{blue}{rollV} &  \textcolor{red}{in}-\textcolor{red}{NP-Loc} & \textsc{Pc-rr}  \\ \\
\end{tabular}

\vspace{0.5cm}
  
\begin{tabular}{llll} 
\hline
\multicolumn{4}{c}{\sc Context} \\
\hline
1 & \textcolor{violet}{The man} & \textcolor{blue}{rolled}  & \textcolor{orange}{the dice} \\
2 & \textcolor{violet}{The man } & \textcolor{black}{did}  & \textcolor{orange}{it}\\

3 & \textcolor{orange}{The dice} & \textcolor{black}{was} & \textcolor{red}{in the cup}\\
4 & \textcolor{orange}{The dice} & \textcolor{blue}{rolled} & \textcolor{red}{in the cup}\\
5 & \textcolor{violet}{The explorer} & \textcolor{blue}{rolled} & \textcolor{orange}{the mat}\\
6 & \textcolor{violet}{The explorer} & \textcolor{black}{did} &\textcolor{orange}{it}\\
7 & \textcolor{orange}{The mat} & \textcolor{black}{was}  & \textcolor{red}{into a pillow}\\ 
8 & ??? 
\end{tabular} 
\hspace{0.4cm}
\begin{tabular}{lp{0.35\textwidth}} 
\hline
\multicolumn{2}{c}{\sc Answers } \\
\hline
1 & The mat rolled the explorer \\
2 & The explorer was into a pillow \\
3 & The explorer rolled into a pillow \\
4 & \textcolor{orange}{The mat} \textcolor{blue}{rolled}   \textcolor{red}{into} \textcolor{red}{a pillow}\\
5 & The dice rolled the man \\
6 & The man was in the cup \\
7 & The man rolled in the cup \\\\
\end{tabular}
    \caption{BLM Roll template and example. Errors labels indicate combinations of paradigm shifts or changes (PS, PC), structure changes (SC), role errors (RR), role swaps (RS).}
    
    \label{fig:blm-roll-eg}
\end{figure}

\subsubsection{The object-drop BLM template}

\begin{figure}[!h]
\centering
\footnotesize 
\setlength{\tabcolsep}{2pt} 

\begin{minipage}{0.49\textwidth}
\begin{tabular}{lllll} 
\hline
\multicolumn{5}{c}{\sc OD context}\\
\hline
1 & \textcolor{violet}{NP-Agent} & \textcolor{brown}{ActiveV} & \textcolor{orange}{NP-Theme} & \textcolor{blue}{P}-NP\\
2 & \textcolor{violet}{NP-Agent} & \textcolor{brown}{ActiveV} & \textcolor{orange}{NP-Theme} & \textcolor{red}{by}-NP \\
3 & \textcolor{orange}{NP-Theme} & \textcolor{teal}{PassiveV} & \textcolor{red}{by}-\textcolor{violet}{NP-Agent}  & \textcolor{blue}{P}-NP\\
4 & \textcolor{orange}{NP-Theme} & \textcolor{teal}{PassiveV} & \textcolor{red}{by}-\textcolor{violet}{NP-Agent}  & \textcolor{red}{by}-NP\\
5 & \textcolor{orange}{NP-Theme} & \textcolor{teal}{PassiveV} & & \textcolor{blue}{P}-NP\\
6 & \textcolor{orange}{NP-Theme} & \textcolor{teal}{PassiveV} & & \textcolor{red}{by}-NP\\
7 & \textcolor{violet}{NP-Agent} & \textcolor{brown}{ActiveV} & & \textcolor{blue}{P}-NP\\
8 & ??? & & &  \\
\end{tabular}
\end{minipage}
\hfill
\begin{minipage}{0.49\textwidth}
    \begin{tabular}{rll} \hline
    \multicolumn{3}{c}{\sc OD answers}  \\ \hline
    1 & \textcolor{orange}{NP-Theme} \textcolor{brown}{ActiveV}   \textcolor{red}{by}-NP & \textsc{I-Int}\\
    2 & \textcolor{violet}{NP-Agent} \textcolor{brown}{ActiveV}   \textcolor{red}{by}-NP & \textsc{Correct}  \\ 
    3 & \textcolor{orange}{NP-Theme} \textcolor{teal}{PassiveV}  \textcolor{red}{by}-\textcolor{violet}{NP-Agent} & \textsc{IER-Pass}\\
    4 & \textcolor{violet}{NP-Agent} \textcolor{teal}{PassiveV}   \textcolor{red}{by}-\textcolor{orange}{NP-Theme} & \textsc{ER-Pass}\\
    5 & \textcolor{orange}{NP-Theme} \textcolor{brown}{ActiveV}   \textcolor{violet}{NP-Agent} & \textsc{IR-Trans}\\
    6 & \textcolor{violet}{NP-Agent} \textcolor{brown}{ActiveV}   \textcolor{orange}{NP-Theme} & \textsc{R-Trans} \\
    7 & \textcolor{orange}{NP-Theme} \textcolor{brown}{ActiveV}   \textcolor{red}{by}-\textcolor{violet}{NP-Agent} & \textsc{IE-WrBy} \\
    8 & \textcolor{violet}{NP-Agent} \textcolor{brown}{ActiveV}   \textcolor{red}{by}-\textcolor{orange}{NP-Theme} & \textsc{E-WrBy}  \\ 
    \end{tabular}
\end{minipage}

\caption{ BLM template for OD contexts and answers.
}
\label{fig:ODtemplateBLM}
\end{figure}

\begin{figure*}[h]
    \footnotesize
    \setlength{\tabcolsep}{0.5mm}
    \begin{tabular}{cc}\\
    \begin{minipage}[t]{0.49\linewidth}
    \begin{tabular}{lp{0.9\linewidth}}
    \hline
    \multicolumn{2}{c}{\sc OD Context} \\
    \hline
    1 & \textcolor{violet}{La zia} \textcolor{brown}{mangia} \textcolor{orange}{una bistecca} \textcolor{blue}{nella} sala grande \\
    2 & \textcolor{violet}{La presidente} \textcolor{brown}{può mangiare} \textcolor{orange}{una bistecca} \textcolor{red}{da} programma \\
    3 & \textcolor{orange}{La specialità della casa} \textcolor{teal}{deve essere mangiata} \textcolor{violet}{dalla turista} \textcolor{blue}{nella} sala grande\\
    4 & \textcolor{orange}{Una bistecca} \textcolor{teal}{fu mangiata} \textcolor{violet}{dalla presidente} \textcolor{red}{da} sola\\
    5 & \textcolor{orange}{La specialità della casa} \textcolor{teal}{deve essere mangiata} \textcolor{blue}{in} un secondo \\
    6 & \textcolor{orange}{Una bistecca} \textcolor{teal}{deve poter essere mangiata} \textcolor{red}{da} sola\\
    7 & \textcolor{violet}{La turista} \textcolor{brown}{deve mangiare} \textcolor{blue}{con} fame \\
    8 & ??? 
    \end{tabular} 
    \end{minipage}
    &
    \begin{minipage}[t]{0.49\linewidth}
    \begin{tabular}{lp{0.9\linewidth}}
    \hline
    \multicolumn{2}{c}{\sc OD Answers} \\
    \hline
    
    1 & La specialità della casa può mangiare da sola \\
    2 & \textcolor{violet}{La squadra di calcio} \textcolor{brown}{deve mangiare} \textcolor{red}{da} mezz'ora\\
    3 & Una bistecca è mangiata dalla turista\\
    4 & La squadra di calcio può essere mangiata da una carbonara\\
    5 & La pasta col pomodoro può mangiare la squadra di calcio\\
    6 & La squadra di calcio mangia una bistecca \\
    7 & La specialità della casa deve poter mangiare dalla turista\\
    8 & La presidente mangia da una bistecca\\
    \end{tabular}
    \end{minipage}
  
    \end{tabular}
    \caption{Example of type II BLM-OD data in Italian.}
    \label{fig:datainstances}
\end{figure*}

In contrast to  the previous causative BLM matrices,  the verbs belonging to the Object-drop alternation also exhibit a transitive/intransitive alternation, but the subject bears the same semantic role (\textsc{Agent}) in both the transitive and intransitive forms and the verb does not have a causative meaning (\textit{The artist paints this door, the artist paints}) \citep{Levin93,merlo2001automatic}. 
The BLM template of Object-drop verbs  differs minimally from the Change-of-state context template, in the intransitive followed by \textcolor{blue}{p}-NP. The minimal difference between the two classes is the semantic attribute of the subject of the intransitive representing the last sentence of the context (sentence 7 in the contexts in Figure \ref{tab:COStemplateBLM}). Consequently, the correct answer also varies across the two groups, but, in both cases, they still structurally represent an intransitive form followed by \textcolor{red}{by}-NP.
The answer set of OD verbs is the same as CoS verbs, even if  the correct answer is not the same. For example, in the English dataset, the correct  answer for COS is an error for OD, while the correct answer for OD is considered a mistake for CoS.\footnote{This is not entirely true for the Italian data, which maintains the morphosyntax of the correct alternation (e.g., the presence of the reflexive-like element).} 
The templates for the context and the answer set for Object-drop verbs are  provided in Figure \ref{fig:ODtemplateBLM}. 
The corresponding examples are provided in Figure \ref{fig:datainstances} (in Italian to illustrate the linguistic variety of the datasets).

\subsubsection{The agreement BLM template}

In subject-verb agreement the primary property is that the subject and the verb must match in grammatical number (singular and plural). A secondary property indicates that these two elements do not need to be linearly adjacent, because the rule applies on the syntactic structure of the sentence and not on its linear order \citep{gulordava-ea18,linzen-ea16}. In this respect, we use intervening elements to indicate that agreement is not a linear property.
The intervening elements, called attractors, can have a confounding effect.

The matrix is organised as a sequence generated by a rule of progression of number of attractors (one and two), a rule of subject-verb agreement that alternates every sentence between singular and plural of the head noun ---to indicate that agreement applies to both singular and plural--- and a rule of number of the attractors that alternates between singular and plural every two sentences. Thus, the correct answer for this example is a sentence that has three noun phrases and a plural subject and plural first attractor and singular second attractor. It is chosen among sentences that violate agreement, or the progression rules or are simply the wrong structure. 
The template and example of BLM for agreement were provided in Figure \ref{fig:template-matrices}, which also explains the logic of the errors in the answer set.

\subsubsection{The sequence-of-tense and mixture BLM template}

The Sequence-of-Tense matrix exemplifies a different kind of long-distance agreement problem, the sequence-of-tenses rule, governing the coherent use of tenses in different sentences. It is especially used in hypotheticals, which require subjunctive and conditional mood, in the present, past and future tenses. The  template is constructed for Italian, so that the verb conjugation is  clearly marked morphologically, both in mood and tense, also making a difference between singular and plural.

The mixture BLM merges both the sequence of tense rule and the agreement rule, which in this case covers both number and gender agreement. The pattern template of the data and an example are shown in section \ref{app:template} in the appendix.
As can be noticed, the agreement pattern is similar to the pattern shown in Figure \ref{fig:template-matrices}, with number agreement alternating at every sentence (Sg, Pl, Sg, Pl,..), the gender agreement pattern alternating every two sentences (M, M, F, F,...) and an intervening attractor, exhibiting no agreement pattern, which is absent in the first four sentences and present in the last four sentences. Notice that, in this matrix, the attractor is an incidental sentence, so it also interferes in structure and distance with the sequence of tenses, by introducing a third verb and lengthening the distance between the elements in the tense sequence.

\subsection{Conclusions}

The abstract templates we develop here fulfil a partly different role from the patterns used to create other grammatical large scale datasets, like BLiMP for example, among others \cite{warstadt-etal-2020-blimp-benchmark}. Like other partially synthetic datasets, the templates are used to generate actual sentences with a process of lexical instantiation (described in the next section). And like minimal contrastive pairs the data generation must obey some basic relation across sentences. 
But the BLM templates are part of the task to be solved themselves. The task solution requires going beyond a binary decision anchored in two contrastive words in the minimal pair. It requires solving a whole system of rules in a coherent way. So the template itself is not an expedient to generate sentences, but it is an integral part of the problem to test linguistic and logical abilities. As such, BLMs are closer to the family of tasks defined by ARC \cite{chollet2019}, rather than to the family of tasks defined by BLiMP.

The development of a BLM template for our linguistic problems is expert-dependent and time-consuming, but it is a core step in developing our rich datasets. 
It is precisely the choice of structure, the way to embody the underlying rules, the choice of interplay between the linguistic and the sequential component of the matrix that makes the originality and structural richness of these datasets, and makes them so hard to beat, compared to other tasks, (as shown in \citet{attanasio-etal-2024-calamita,nissim-ea2025}), providing a strong test for linguistic generalisation.

\section{The BLM Datasets in Many Languages}
\label{BLM-descriptions}

\label{blmgeneration}

\begin{figure}
    \centering
    \includegraphics[width=0.8\linewidth]{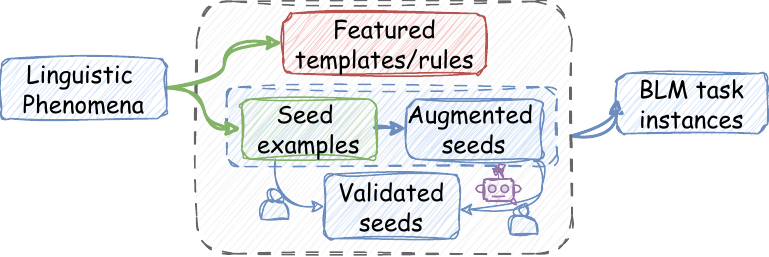}
    \caption{Data flow for the automatic creation of the BLM structured datasets.}
    \label{pipeline}
\end{figure}

The abstract templates described in the previous section provide a pattern for the creation of the actual datasets. 
From the linguistic phenomenon to the creation of the lexical seed set, various approaches can be pursued based on the type of linguistic phenomenon being investigated, using natural occurring examples or experimental stimuli or synthetic data. As will be articulated in the next subsections, this choice might depend on whether the phenomenon has already been extensively studied in experimental linguistics, the scale of the lexical components involved in the linguistic phenomenon, and the available resources in the target language.  
Figure \ref{pipeline} summarizes the pipeline that is followed for the whole process and that will be described in more detail in the next sections. 

\subsection{The \textit{spray/load} dataset in English (s/lE)}

\paragraph{Developing lexical seeds}

Dedicated rich repositories exist that group verbs into semantically coherent classes based on shared syntactic behavior \citep{stowe-etal-2021-semlink}. We retrieved the lexical verbs belonging to the same class as \textit{spray} and the verb \textit{load} from \textsc{VerbNet} (\citealt{Kipper2005})\footnote{\url{https://verbs.colorado.edu/verb-index/vn/spray-9.7.php\#spray-9.7-1}.} to identify a set of 30 verbs that undergo the \textit{spray/load} alternation.\footnote{In alphabetical order: \textit{baste, brush, drizzle, hang, load, plaster, pump,  rub,   scatter, seed, sew,  shower,  smear,  smudge, sow,  spatter,  splash, splatter, spray, spread, sprinkle, spritz, spurt, squirt, stick, strew, string, swab, swash,  wrap.}}
The data generation process begins by manually selecting a naturally occurring example for each verb in the class, from the SPIKE platform \citep{shlain-etal-2020-syntactic}.\footnote{\url{https://spike.apps.allenai.org/datasets}} SPIKE offers a collection of English corpora from various text genres (encyclopedic entries, internet reviews, among others). We favoured  natural contexts, and  examples obtained from the Amazon Reviews subcorpus. We performed the search of different inflected forms of the verbs and analysed the naturally occurring examples in the output manually.\footnote{
For example, the search for different inflected forms of the verb \textit{spritz} allowed us to retrieve a sentence like \textit{I spritz the fish with low-fat oil} (Amazon Customer; A19VW07IUHNV1R).
We removed non-target examples (e.g. those cases in which the PP does not represent an argument of the verb, but it is  embedded in the nominal phrase, as in \textit{This bulb scatters dim light with a green hue around}; ID: AMG Enthusiast; AI9JTX5ZX0OJG).} Care was also taken to vary the types of sentences as much as possible (e.g. imperatives, relatives) and pronominal entities. These construction choices counter the fact that a good portion of Amazon reviews involving these verbs are in first person singular.

\paragraph{Masked augmentation} The extracted examples were then used to generate more examples. Each of the three arguments 
(\textsc{Agent}, \textsc{Theme} and \textsc{Loc}) was masked in two contexts, one aiming at eliciting an indefinite noun (without the article, e.g. \textit{I spritz [MASK] with low fat oil}) and the other at eliciting a definite noun (preceded by the definite article \textit{the}, eg. \textit{I spritz \textbf{the} [MASK] with low fat oil}).  We used the DistilBERT uncased model \citep{distilbert}, as we want to use a different language model for data augmentation from the one we will use for learning,  to avoid bias.  From the proposed items, we manually selected five elements that, when combined, produced grammatically and semantically acceptable sentences (for example, they belonged to a similar semantic field as the seed). We also took definiteness into account. Pronouns were limited to those present in naturally occurring examples. 
Two of the authors individually validated each item. In the end, each of the 30 verbs was related to five \textsc{agents}, five \textsc{theme} and five \textsc{loc}. We refer to this group of sentences as the \textit{lexical seed set}.

\paragraph{Instantiation and lexical variants}

The constituents belonging to the lexical seed set were merged to build templates and answer sets. We constructed three types of contexts and answer sets, each of increasing lexical variation.
Type I refers to problems generated directly with seed input segments and their variations: the same lexical elements for the verb, \textsc{Agent}, \textsc{Theme}, and \textsc{Loc} are found for every sentence in each context for a given BLM.  In Type II, a BLM template is built with the same verb, but the arguments, \textsc{Agent}, \textsc{Theme}, and \textsc{Loc}, vary lexically. Finally, a BLM of Type III results from a fully random reshuffle of Type II BLMs, so that all lexical elements vary.

\subsection{The causative dataset (CausE/I)}

\paragraph{Developing lexical seeds}
We selected 30 verbs from each the two classes Change of State and Object-drop discussed in \citet{Levin93}. 
The data is developed both for English and Italian, so the choice of the classes depends on whether both English and Italian maintain the same structure. 
\footnote{All the Change-of-state verbs are for English:\textit{bake, bend, blacken, break, brighten, caramelize}, \textit{chip}, \textit{close}, \textit{corrode}, \textit{crinkle}, \textit{defrost}, \textit{empty}, \textit{expand}, \textit{fry}, \textit{harden}, \textit{harmonize}, \textit{heat}, \textit{improve}, \textit{increase}, \textit{intensify}, \textit{melt}, \textit{open}, \textit{propagate}, \textit{purify}, \textit{sharpen}, \textit{shrink}, \textit{sweeten}, \textit{tear}, \textit{whiten}, \textit{widen}. Italian: \textit{addolcire}, \textit{affilare}, \textit{allargare}, \textit{annerire}, \textit{aprire}, \textit{armonizzare}, \textit{caramellare}, \textit{chiudere}, \textit{corrodere}, \textit{cuocere}, \textit{espandere}, \textit{friggere}, \textit{illuminare}, \textit{indurire}, \textit{ingrandire}, \textit{intensificare}, \textit{migliorare}, \textit{piegare}, \textit{propagare}, \textit{purificare}, \textit{riscaldare}, \textit{rimpicciolire}, \textit{rompere}, \textit{scheggiare}, \textit{sciogliere}, \textit{scongelare}, \textit{sbiancare}, \textit{stropicciare}, \textit{svuotare}. 

The Object-drop verbs are for English: \textit{clean}, \textit{cook}, \textit{draw}, \textit{drink}, \textit{eat}, \textit{fish}, \textit{hum}, \textit{iron}, \textit{knead}, \textit{knit}, \textit{mend}, \textit{milk}, \textit{nurse}, \textit{paint}, \textit{play}, \textit{plow}, \textit{polish}, \textit{read}, \textit{recite}, \textit{sculpt}, \textit{sew}, \textit{sing}, \textit{sow}, \textit{study}, \textit{sweep}, \textit{teach}, \textit{wash}, \textit{weave}, \textit{whittle}, \textit{write}. Italian: \textit{allattare}, \textit{arare}, \textit{bere}, \textit{cantare}, \textit{canticchiare}, \textit{cucire}, \textit{cucinare}, \textit{dipingere}, \textit{disegnare}, \textit{impastare}, \textit{insegnare}, \textit{intagliare}, \textit{lavare}, \textit{leggere}, \textit{lucidare}, \textit{mangiare}, \textit{mungere}, \textit{pescare}, \textit{giocare}, \textit{rammendare}, \textit{recitare}, \textit{saldare}, \textit{scolpire}, \textit{seminare}, \textit{spazzare}, \textit{stirare}, \textit{studiare}, \textit{tessere}, \textit{scrivere}.}
\paragraph{Masked Augmentation} The creation of the lexical sentences starts from the English data. These data were manually selected from a list of the top 25 results automatically generated using masked modelling.\footnote{The language model used was a different language model (\textit{bert-base-uncased}, \citealt{devlin-etal-2019-bert}) to avoid any form of bias.} Specifically, our input for the masking were the lexical verbs only surrounded by functional elements to retrieve suitable agents and patients. For example, to retrieve the arguments for the verb \textit{break}, we masked the patient leaving the agent in a pronominal form (e.g. \textit{she broke (the/a/some/...) [MASK]}), and masked the subject of the transitive leaving the patient pronominal to retrieve plausible agents \textit{(e.g. (the/a/some/...) [MASK] broke it}. The Italian data are derived from the English, with manual intervention and validation from native speakers, when required to guarantee the acceptability of the sentences, and semantic plausibility.

\paragraph{Instantiation and lexical variants}
Each dataset contains 9000 instances of BLMs (3000 for each type), semi-automatically crafted and manually evaluated for plausibility and grammaticality by two (near-)native speakers.

\subsection{The \textit{ Roll} verbs dataset in English (BLM-RollE)}
\paragraph{Developing lexical seeds}
We selected verbs from Levin's \textit{Roll}-class \citep{Levin93}. The class comprises verbs expressing dynamic motion \footnote{ \textit{bounce, drift, drop, float, glide, move, roll, slide, swing}} and verbs denoting motion around an axis. \footnote{\textit{coil, revolve, rotate, spin, turn, twirl, twist, whirl, wind}.}
We manually crafted 64 minimal sentence pairs covering both transitive  and intransitive structures, with prepositional adjuncts (PP) for naturalness.
These seed pairs represent the argument structure alternation where the same entity appears as subject in the intransitive (\textit{The ball rolled into the corner}) and as object in the transitive (\textit{The player rolled the ball into the corner}).

\paragraph{Masked augmentation}

To increase the diversity in the  dataset, we applied two complementary augmentation methods. Elements requiring variation were identified based on the linguistic alternation properties---specifically, agents, themes, and locative adjuncts that preserve the structural and semantic properties of the causative/inchoative relationship.  In the first augmentation method, using pre-trained RoBERTa-Large \cite{liu2019roberta}, we generated candidate replacements for masked positions, selecting the top five most probable alternatives based on predicted scores.
In  the second augmentation method, we used ChatGPT (GPT-4) with Python code snippets as prompts to ensure consistency and reduce ambiguity compared to natural language prompts. The process included structured testing phases to verify code comprehension and output quality before large-scale generation.
Both approaches augmented the seed sentences with semantically coherent alternatives while maintaining the core argument structure relationships required for the causative/inchoative alternation.

\paragraph{Instantiation and lexical variants}
Following augmentation, all generated sentences were validated by two human linguistic experts and ChatGPT using 5-point Likert scales for grammatical correctness and semantic plausibility. 
%
The dataset generates two lexical variants: Type I has identical verbs within contexts, and Type II has two distinct verbs but maintains the same
lexical distribution.

\subsection{The Agreement datasets in English and Romance languages: Agr (E,F,I,R) }

\paragraph{Developing lexical seeds}
The seed data for French was created by manually completing previously published data \cite{franck2002subject}. 
Each subset contains three clause structures (main, relative, completive)  uniformly distributed within the data. The dataset separates sequence-based from other types of errors, to be able to perform deeper analyses into the behaviour of pretrained language models. 
From this initial data, we generated a dataset that comprises three subsets of increasing lexical complexity (as described above, details in \citet{an-etal-2023-blm}): Types I, II, III,  corresponding to different amounts of lexical variation within a BLM problem instance. 

\paragraph{Masked Augmentation and lexical variants}
The set produced by applying the seed sentences is  dataset type I.
To introduce some lexical variation in this dataset in a semi-automatic manner,  CamemBERT \cite{martin-etal-2020-camembert} is used to replace individual words in the sentences in the type I dataset. We call this dataset type II.
To further increase the lexical variation, be build the type III dataset, where a BLM problem consists of a combination of sentences (with the same grammatical structure) from different type II problems.

\paragraph{Multilingual data creation} The datasets in English, Italian and Romanian were created by manually translating the seed French sentences into the other languages by native (Italian and Romanian) and near-native (English) speakers. The internal structure in these languages is very similar, so translations are approximately parallel. The differences lie in the treatment of preposition and determiner sequences that must be conflated into one word in some cases in Italian and French, but not in English. 
French and Italian use number-specific determiners and inflections, while Romanian and English encode grammatical number exclusively through inflections. In English most plural forms are marked by a suffix. Romanian has more variation, and noun inflections also encode case. Determiners are separate tokens in English, Italian and French, and in Italian and French they are also overt indicators of grammatical number and of phrase boundaries, whereas inflections may or may not be tokenized separately. 

For augmenting the French data, first the words for which alternatives are sought are chosen (verbs and nouns in the sentence). We then use a masked language model and mask each word in turn. The highest probability seven options for each word are written to a json file, which is manually checked and edited. Type II variation data are generated from this, and the type III by mixing the type II problems based on the sentence patterns. For the additional languages (English, Italian, Romanian) the seeds are created manually by a native speaker starting from the French data (not always translations of the French version, but we tried to stay close). The augmentation process is the same. We have used multilingual BERT.\footnote{\url{https://huggingface.co/google-bert/bert-base-multilingual-uncased} to obtain lexical alternatives for all four languages. }

\subsection{Data Validation}

As mentioned already, the created data is validated at different stages and with different methods. The lexical seeds, which are often derived from naturalistic corpora or controlled experimental stimuli, are always also manually validated by more than one author.  A fill-mask procedure with transformers or a Large Language Model is then used to automatically generate additional, plausible constituents for the desired structures.  These created combinations are also validated manually by the authors.

The evaluation of linguistic outputs in terms of their grammatical and semantic plausibility has a long-standing tradition in linguistic studies. Grammatical correctness and semantic coherence are interrelated, but distinct, dimensions of sentence plausibility \cite{schutze2013judgment}. While the formal rules governing sentence structure are assessed through grammatical correctness, semantic plausibility depends on the meaningfulness and coherence of a sentence in its given context~\cite{sprouse2018colorless}. Recent NLP analyses underline the separate distributional behaviours exhibited by these criteria~\cite{sprouse2018colorless,warstadt2018cola}. As models become adept at producing syntactically correct outputs, the nuances of semantic correctness become an important criterion in evaluating model quality. 
A spot-checking experiment of human validation to evaluate the grammatical correctness and semantic plausibility of the sentences created by these processes was  run on the \textit{Roll}-verbs dataset.

Some of the authors, proficient speakers of English with advanced linguistic training, but who had not created  the data, performed the validation. For the  reasons indicated above, we opted for separate evaluations for grammaticality and plausibility.
To avoid overload and consequent mistakes, the datasets were partitioned into ten questionnaires, each containing 192 sentences. Each validator saw all questionnaires. This partitioning distributes the workload over time and maintains the quality of annotations, and consistency in validation judgements. Both types of plausibility were evaluated (see appendix for instructions). During validation, the annotators were presented with individual sentences on a screen with two distinct queries about grammatical and semantic plausibility, using a sliding Likert scale from $1$ to $5$ for scoring, where $1$ represents ``strongly unacceptable'' and $5$ represents ``strongly acceptable''.
(See Figure \ref{fig:likert-validation} in the appendix). 
We used PsyToolkit \cite{stoet17} for data collection.

\begin{table}[ht!]
\centering
\small
\resizebox{0.9\textwidth}{!}{%
\begin{tabular}{llllll}
\hline
\multirow{2}{*}{\textbf{Model}} &
  \multirow{2}{*}{\textbf{Structure}} &
  \multicolumn{2}{c}{\textbf{Grammatical   plausibility}} &
  \multicolumn{2}{c}{\textbf{Semantic plausibility}} \\ \cline{3-6} 
                         &         & \textbf{human\_a} & \textbf{human\_b} & \textbf{human\_a} & \textbf{human\_b} \\ \hline
\multirow{2}{*}{GPT-4}   & intrans & $4.37\pm0.72$     & $4.68\pm0.68$     & $4.11\pm1.02$     & $4.12\pm1.45$        \\
                         & trans   & $4.74\pm0.52$     & $4.75\pm0.56$     & $4.54\pm0.83$     & $4.47\pm1.15$     \\ \hline
\multirow{2}{*}{RoBERTa} & intrans & $4.30\pm0.77$     & $4.44\pm0.90$     & $3.98\pm1.15$     & $3.55\pm1.75$       \\
                         & trans   & $4.68\pm0.60$     & $4.61\pm0.79$     & $4.39\pm1.01$     & $4.19\pm1.44$        \\ \hline
\end{tabular}%
}
\caption{Descriptive statistics (mean $\pm$ {standard deviation}) for grammatical and semantic plausibility scores, categorised by model and sentence structure, as rated by different evaluators.}
\label{tab:descriptif-stat}
\end{table}

\begin{table}[]
\setlength{\tabcolsep}{1mm}
\centering
\small
\begin{tabular}{l|lcccc}
\hline
\textbf{Model} & \multicolumn{1}{c}{\bf Sentence} & \multicolumn{2}{c}{\textbf{GP}} & \multicolumn{2}{c}{\textbf{SP}} \\ \cline{3-6} 
                         & \multicolumn{1}{c}{}                       & h$_a$ & h$_b$                 & h$_a$ & h$_b$ \\ \hline
&\multicolumn{5}{c}{\textit{Sentences with unanimously high ratings}}                                        \\ \hline
{GPT-4}   & The gambler dropped the dice.     & 5     & \multicolumn{1}{c|}{5} & 5     & 5     \\
          & The handle turned to the left.    & 5     & \multicolumn{1}{c|}{5} & 5     & 5     \\ \cline{2-6} 
{RoBERTa} & The worker moved the machine.     & 5     & \multicolumn{1}{c|}{5} & 5     & 5     \\
                         & The phone dropped into the water. & 5     & \multicolumn{1}{c|}{5} & 5     & 5     \\ \hline
& \multicolumn{5}{c}{\textit{Sentences with discrepancies in ratings}}                                         \\ \hline
{GPT-4}   & The student revolved the pencil.  & 4     & \multicolumn{1}{c|}{2} & 4     & 2  \\
                         & The tenant moved into storage.    & 5     & \multicolumn{1}{c|}{4} & 5     & 2  \\
                         & The bottle twisted on the finger. & 4     & \multicolumn{1}{c|}{2} & 2     & 1  \\
                         & The camper glided the kayak.      & 5     & \multicolumn{1}{c|}{4} & 2     & 1  \\ \cline{2-6} 
{RoBERTa} & The golfer swung the driver.                  & 5  &\multicolumn{1}{c|}{5} & 5         & 1       \\
\multicolumn{1}{l|}{}    & The baby wound around the ankle.  & 3     & \multicolumn{1}{c|}{1} & 5     & 1   \\
\multicolumn{1}{l|}{}    & The numbers rolled into a pillow. & 5     & \multicolumn{1}{c|}{5} & 1     & 1   \\
\multicolumn{1}{l|}{}    & The phone twisted on the finger.  & 5     & \multicolumn{1}{c|}{4} & 1     & 1   \\ \hline
\end{tabular}%
\caption{Example of sentences generated with GPT-4 and RoBERTa, with grammatical (GP) and semantic plausibility (SP) judgments, from human judges.}
\label{tab:sents-with-ratings}
\end{table}

Table \ref{tab:descriptif-stat} presents descriptive statistics for the grammatical and semantic plausibility scores attributed to the generated sentences.
Most sentences from both generation methods (Masked and generative use of LLMs) receive relatively high ratings (above $4.0$ on average), which indicates that either approach can produce coherent sentences. The small gap in performance shows that \texttt{RoBERTa} fill-mask remains quite effective, and ChatGPT zero-shot generation produces sentences that are marginally more acceptable.
Overall, these strong ratings indicate that our synthetic generation pipeline can produce controllable and satisfactory sentences. 

\subsection{Conclusions}

We have created a large body of BLM problems and thousands of matrices, with two different generation methods, whose seed sets have all been manually validated, covering seven different types of BLMs with two different internal structures, and spanning four languages.\footnote{The data examples for Sequence of tense and Mixture are much smaller, as they were used only in a few-shot experiments on ChatGPT. They were generated automatically by prompting alternatives and manual validation. } This large, valid and richly-structured dataset can be used to study foundational questions in how LLM process language. The following section demonstrates the usefulness of this data-centric approach by exploring the three core problems of linguistic object identification, structure detection and  systematic learning.  We hope these examples will inspire other studies using BLMs.

\section{Exploring the use and usefulness of BLMs: Three questions on the structure of inner representations}
\begin{figure}
\centering
\includegraphics[scale=0.42]{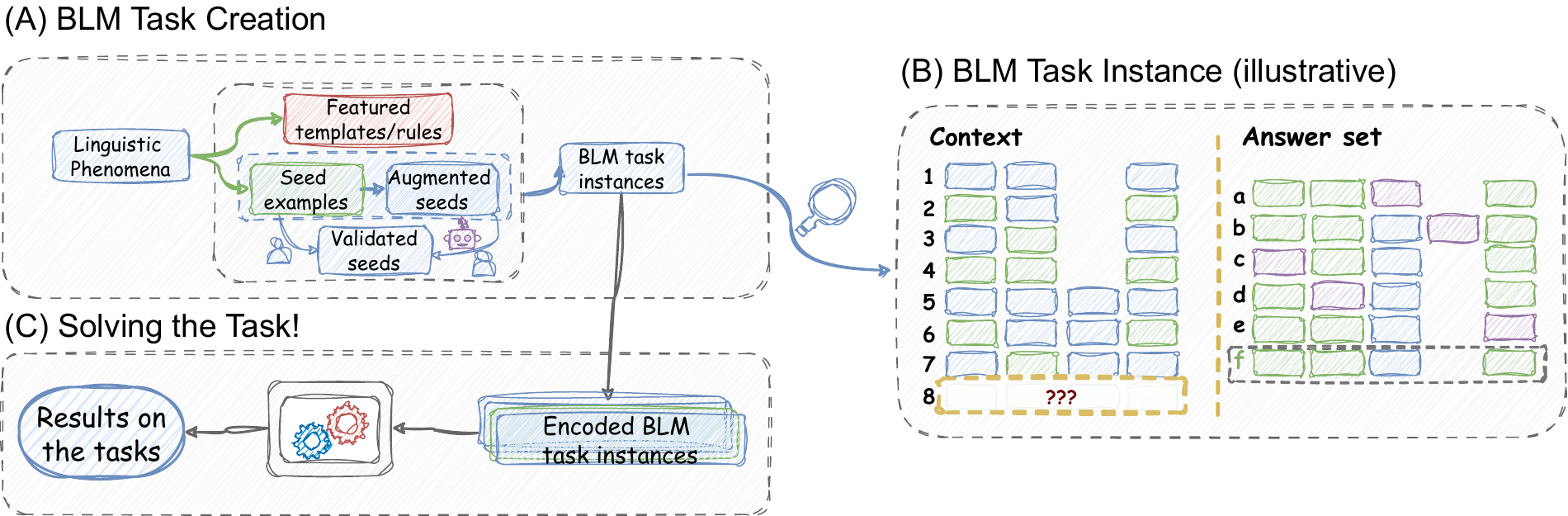}
\caption{Illustration of the flow of BLM template and data creation and the setup for the experiments.}
\label{fig:baselines}
\end{figure}

BLM problems give rise to richly structured datasets.  They  encode a linguistic phenomenon across a sequence of input sentences organised in a structured template; the template acts as an implicit annotation of the underlying linguistic properties and rules; each individual sentence, both in the context and in the answer set,  has structure.  This multi-level structure provides a coherent framework to investigate the inner workings of LLMs in a systematic way. 

Here, we run the experiments that study the core usefulness of these different aspects of the datasets. 
Our experiments are inspired by the way people solve the analogous RPM problem. It has been shown that people start by first identifying the relevant 'objects' and their attributes and then use these components to identify the rules that connect them together \cite{carpenter-ea1990}. 
We organise our experiments in an analogous way: first, we ask \textit{if} the BLM problems can be solved, then we ask \textit{how} the problems are solved, with particular attention to how the linguistic objects and attributes are identified and manipulated, and \textit{if and how} the sequential structure of the templates are learnt.  We visualise the steps in Figure \ref{fig:baselines}. More precisely,  our investigations cluster around the following questions.

\noindent
\paragraph{Q1 Can BLMs be solved?} At what level of accuracy? What are the factors that determine the different levels of accuracy? 
We develop baseline models (CNN and FFNN) and answer positively (section 6). However, the performance scores and error analyses show that the task is challenging, as we expected, and therefore that there is room for improvement.

\paragraph{Q2 Can the internal structure of sentences in a BLM task be detected? } Answers to this question tell us if the linguistic objects on which the BLMs are built can be identified and used to create a solution. In Section \ref{sec:sentences}, we test whether such linguistic objects -- in particular chunks -- can be detected and captured using a VAE-based system, and show supporting evidence in term of results on the task and visualisations.

\paragraph{Q3 Can the systematic structure of BLMs be detected?}
Using VAEs, the input sequence is reduced to an inner, much compressed, representation.
We rely on this compression to identify  the systematic structural relation across the sentences and their sequence in the input (Section \ref{sec:systematicity}).

\section{Assessing the BLM challenge  and its implications}
\label{sec:benchmarking}

The first question to ask is whether the BLM task is solvable, under what conditions and to which levels of performance. For the BLM task to be useful in investigations of representations and learning, it needs to be challenging, but also be solvable at a level of reasonable performance. For any performance and error analysis or any finer-grained analysis to be meaningful, it must be the case that the system is able to address the task. 

\subsection{The two baselines} 

\paragraph{Sentence representations}

We obtain sentence representations as averaged token embeddings \citep{nikolaev-pado-2023-investigating} from an encoder and a decoder model with comparable parameters. The encoder model is the Electra pretrained model \cite{clark2020electra}\footnote{{\it google/electra-base-discriminator}}. We choose Electra because it has been shown to perform better than models from the BERT family on the Holmes benchmark\footnote{The HOLMES benchmark leaderboard: \url{https://holmes-leaderboard.streamlit.app/}. At the time of writing, the ranks were as follows: Electra:16, DeBERTa:21, BERT:41, RoBERTa:45.}, and to also encode information about syntactic and argument structure better \cite{yi-etal-2022-probing,nastase-merlo-2024-identifiable}. It also outperforms XLNet, a variation on BERT with autoregressive training \cite{yang2019xlnet}, and MPNet, a variation of BERT with masking and permuting \cite{song2020mpnet}. The decoder model is OpenAI-GPT \cite{radford-etal-2018-improving}\footnote{\it https://huggingface.co/openai-community/openai-gpt}. It has high performance on benchmarks on textual entailment, semantic similarity, reading comprehension even in zero-shot testing, and compares well with larger sized models despite its relatively low size \footnote{https://huggingface.co/spaces/mteb/leaderboard}.


\paragraph{Baseline Model}
We use a feed-forward neural network (FFNN) as a baseline model.\footnote{In previous work, we also experimented with a Convolutional Neural Network (CNN). Because the sentence embedding produced by the transformer captures structural information and we are presenting sentences in a sequence, both the FFNN and the CNN will have the chance to find patterns shared across the sentences.  The input to the CNN is an array of embeddings, of size 7 x 768.  This is passed through three successive layers of 2-dimensional convolutions, with a kernel size 3x3 (stride 1, no dilation). The output of the convolution is passed through a fully connected layer to compress it to the sentence representation size (768). Because of the kernel size, stride=1, and no dilation, this setup will focus on finding localized patterns in the sentence sequence array. 
The output of the CNN networks is the same -- a vector representing the embedding of the predicted answer. 
We refer to previous publications for results on CNN on Electra embeddings. Overall, the picture painted by the FFNN and CNN baselines is very similar.}

The input to the FFNN is a concatenation of the sentence embeddings in the sequence (size 7 * 768), that is passed through 3 fully connected layers that gradually compress the input (7 * 768 $\xrightarrow{layer 1}$ 3.5 * 768 $\xrightarrow{layer 2}$ 3.5 * 768 $\xrightarrow{layer 3}$ 768) to the size of a sentence representation. Because of the full connectedness between successive layers, the FFNN has the capacity of capturing patterns spread out over the entire input vector. 
The output of the FFNN networks is a vector representing the embedding of the predicted answer. 

The learning objective maximizes the probability of the correct answer from the candidate answer set. Because the incorrect answers in this set are specifically designed to be minimally contrastive from the correct one, we implement the objective through a max-margin loss function. This function combines the distances between the predicted answer and the correct and the wrong ones. We first compute a score for the embedding $e_i$ of each candidate answer $a_i$ in the answer set $\mathcal{A}$ with respect to the predicted sentence embedding $e_{pred}$ as the dot product between the respective vectors:
$
  score(e_i, e_{pred}) = e_i^Te_{pred}
$
The loss uses the maximum margin between the score for the correct answer $e_c$ and for each of the incorrect answers $e_i$, as in (1). At prediction time, we take the answer with the highest $score$ value from a candidate set as the correct answer.\footnote{All systems used a learning rate of 0.001 and Adam optimizer, and batch size 100. The training was done for 120 epochs. The experiments were run on an HP PAIR Workstation Z4 G4 MT, with435 an Intel Xeon W-2255 processor, 64G RAM, and a MSI GeForce RTX 3090 VENTUS 3X OC 24G GDDR6X GPU.}
\begin{equation}
loss_a = \sum_{e_i} [1 - score(e_c, e_{pred}) + score(e_i,e_{pred})]^{+}
\end{equation}

\paragraph{Data statistics}

Table \ref{tab:multilingual-stats} shows the datasets statistics for some of BLM  problems discussed in this paper. 
As a reminder, Type I data is lexically consistent -- the same vocabulary is used in all sentences in the sequence, and in the answer candidates. Type II has a limited amount of lexical variation -- one word in each sentence is different. Type III is more lexically varied, with little, if any, lexical overlap between any of the context or answer candidate sentences. 
After splitting each subset 90:10 into train:test subsets, we randomly sample 2000 instances as train data. 20\% of the train data is used for development. 

\begin{table}
\small
\setlength{\tabcolsep}{2mm}
\begin{tabular}{lcccccccccc}
         & \multicolumn{6}{c}{Verb Alternations}   & \multicolumn{4}{c}{Agreement}  \\
         & \multicolumn{2}{c}{CoS} & \multicolumn{2}{c}{OD}  &\multicolumn{2}{c}{Spray-load} & \multicolumn{4}{c}{ }   \\
         & En & It & En & It & \tiny{(ALT-ATL)} & \tiny{(ATL-ALT)}& English & French & Italian & Romanian  \\ \hline
Type I   &   300   &  300    &  300    &  300    & 230 &  252 &  256 & 256 & 256 & 256  \\
Type II                     &   300   &  300    &  300    &  300   & 1500 & 1500  & 4780 & 4060  & 4830 & 4580  \\
Type III                   &   300   &  300    &  300    &  300   &  1500 & 1500  & 4700 & 4060 & 4830 & 4580  \\\\
\end{tabular}
\caption{Test data statistics for BLM spray/load, BLM-Cos, BLM-OD and BLM-Agr. The amount of training data is always 2000 instances.}
\label{tab:multilingual-stats}
\vspace{-5mm}
\end{table}

\begin{figure}
    \centering
    \includegraphics[width=0.7\linewidth,trim="0 0 0 1cm",clip]{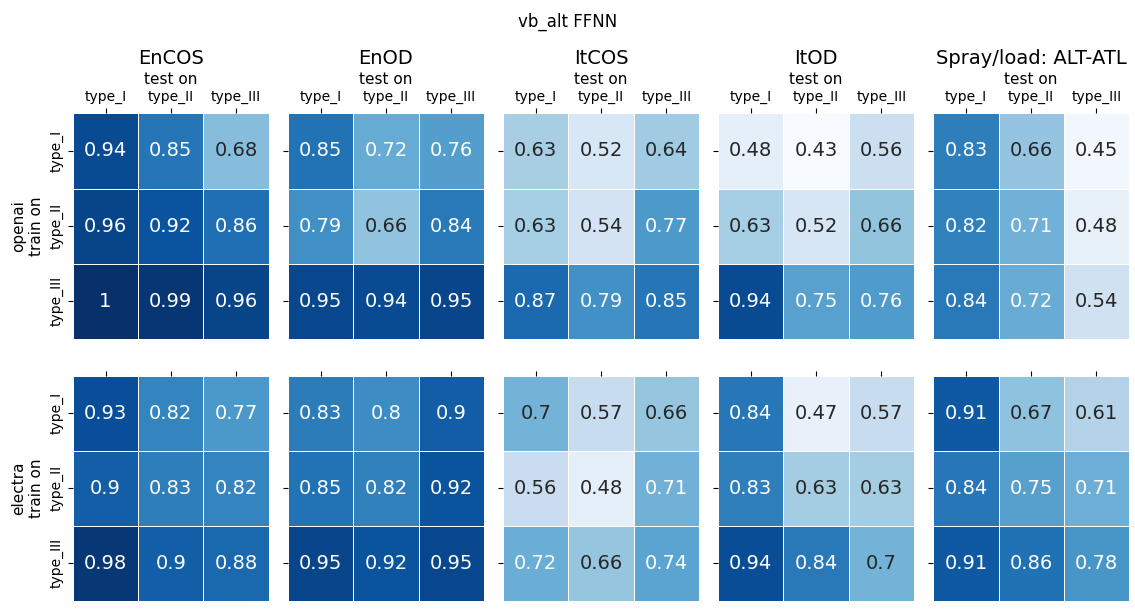}
    \caption{F1 verb alternations Baseline (FFNN with two sentence embeddings): CoS (En,It), OD (En,It), spray/load (En).} \label{fig:baseline_resultsValt}

\end{figure}

\begin{figure}[h]
    \centering
 \includegraphics[width=0.7\linewidth,trim="0 0 0 1cm",clip]{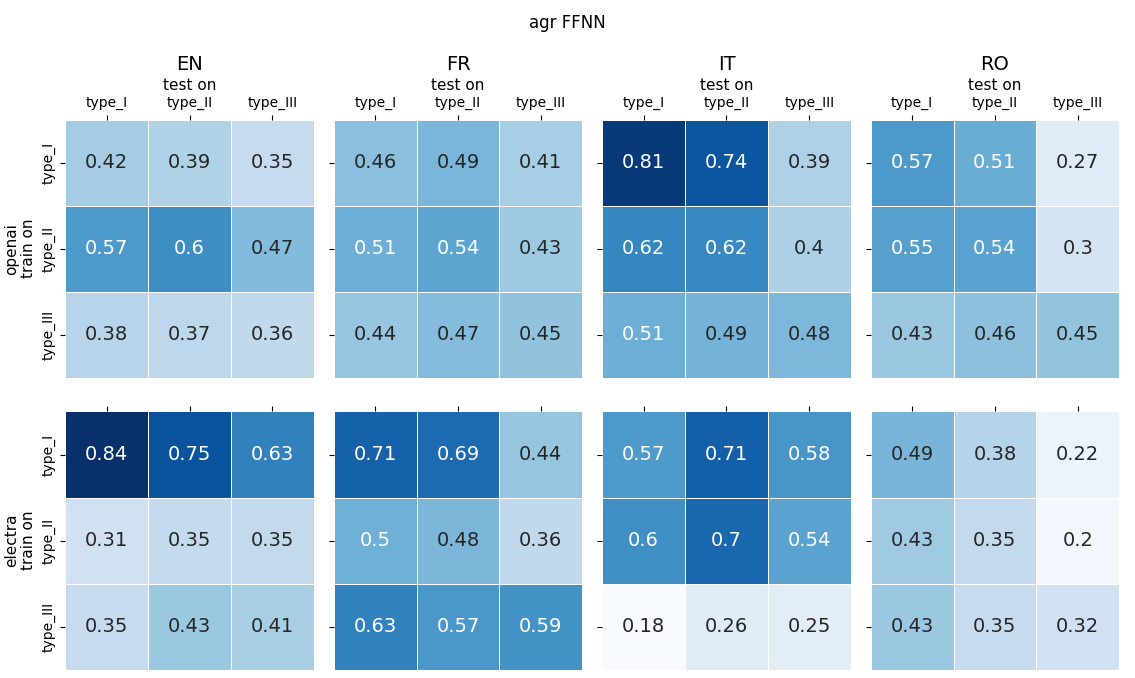}
    \caption{F1 Agr Baselines (FFNN and Electra/Openai sentence embeddings) for 4 languages}
    \label{fig:agrFFNN}
\end{figure}

\subsection{Performance experiments}

\paragraph{Performance in terms of F1 scores}
We report results in terms of average F1 scores -- how well does the model predict the correct answer -- over three runs.
Figures \ref{fig:baseline_resultsValt} and \ref{fig:agrFFNN} show the performance as heatmaps of average F1 scores for the \textit{spray/load} alternation and the CoS and OD alternations and the agreement BLM, in most of the different languages for which they have been developed.
We show the results for the two sentence embeddings. While there are local differences in the results,  the macro-trends across data types (I/II/III) and linguistic problems are similar.

\begin{figure}
    \centering
\includegraphics[width=0.7\linewidth]{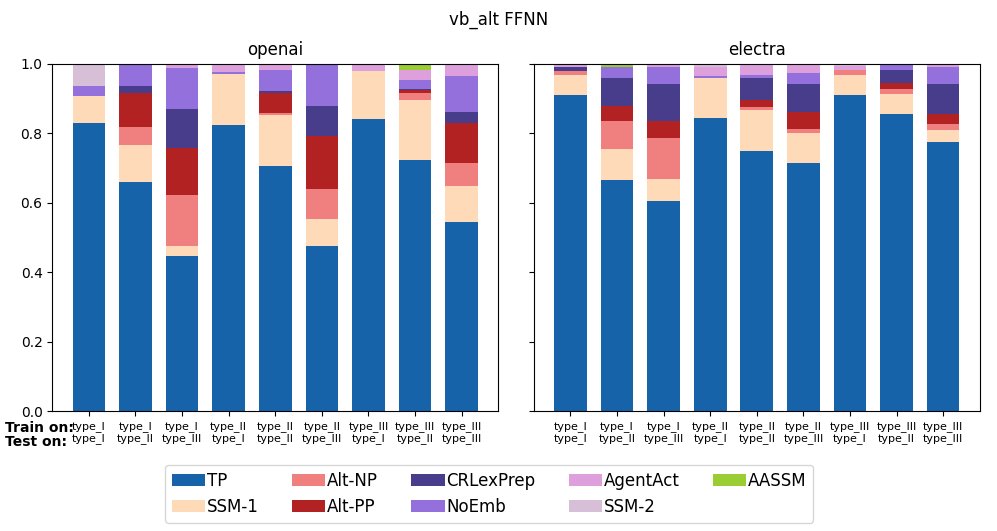}
    \caption{Stacked bars for answer type (correct + errors) distribution for \textit{spray/load} alternation}
    \label{fig:vb_alt_baselines_stacked}
\end{figure}

\begin{figure}
\centering
    \includegraphics[width=0.8\linewidth]{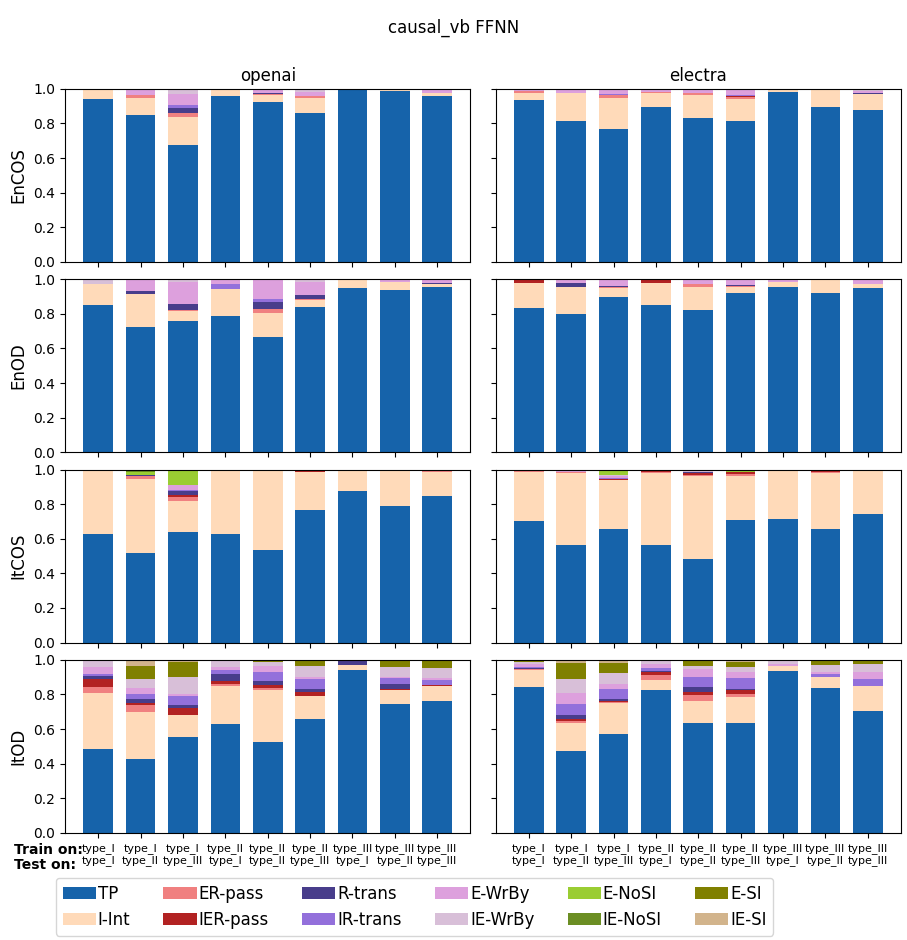}   
    \caption{Stacked bars for answer type (correct + errors) distribution for Change-of-State and Object-drop verbs in English and Italian.}
    \label{fig:cause_vb_baselines_stacked}
\end{figure}

\begin{figure}
\centering
    \includegraphics[width=0.8\linewidth]{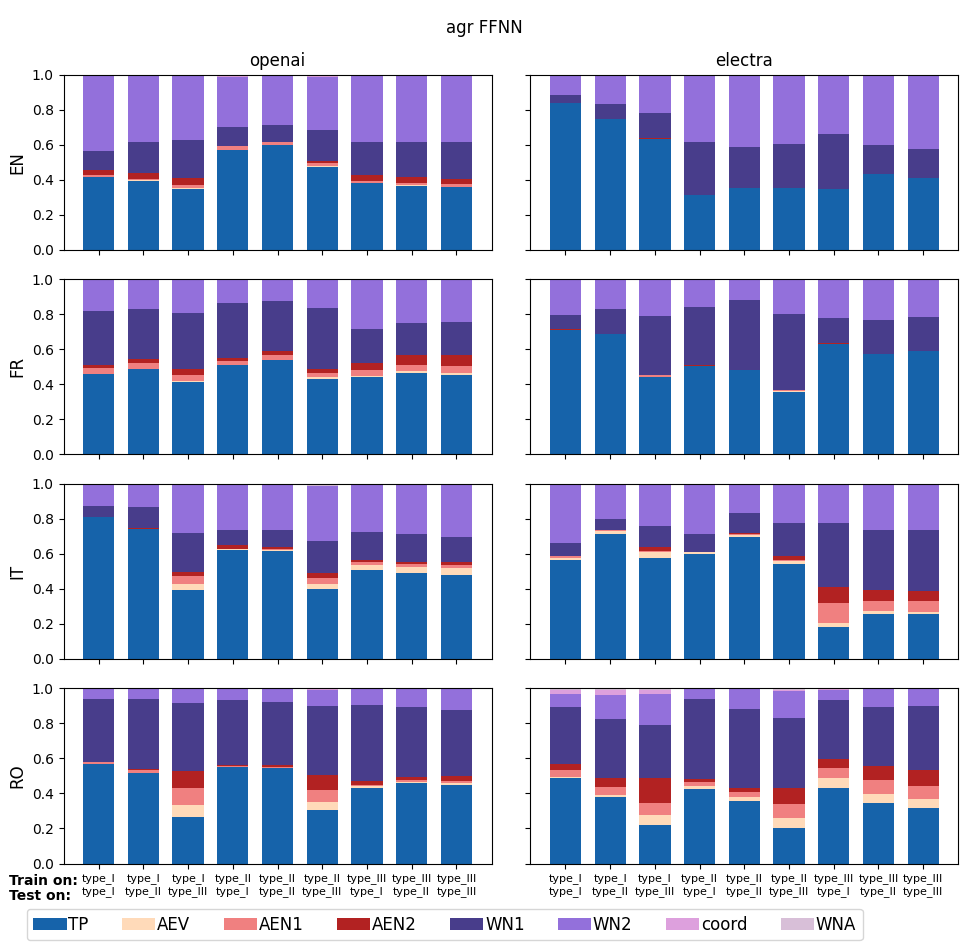}
    \caption{Stacked bars for answer type (correct + errors) distribution for agreement}
    \label{fig:agr_baselines_stacked}
\end{figure}


\noindent \textit{Spray/load} While the results for all configurations are quite high, the most difficult setup is when the model is trained on data with the least amount of lexical variation (type I), and tested in the data with the highest lexical variation (type III). However, even this set-up has a performance of F1 $>$ 0.40, showing that lexical variation is not a deterring factor neither in finding relevant patterns for verb alternations, nor in applying them.

\noindent \textit{CoS/OD alternations}
A similar pattern can be observed for Change of state and object-drop verbs in English and Italian, with two exceptions. First, training on type I and testing on type III shows a marked and unexpected improvement. Second, performance on the Italian data, while it follows the same trends, is overall  lower.

\noindent \textit{Agreement} The FFNN baseline model performs acceptably for English, French and Italian, less so for Romanian. Learning the model based on the type I data -- with the minimum lexical variation -- leads to best results on many set-ups, even when testing on the other subsets, especially for Electra. Having more lexical variation sometimes leads the system to learn a robust model, but more often than not it overwhelms the system, especially for those languages whose embeddings are trained on less data.

\paragraph{Errors}

Figures \ref{fig:vb_alt_baselines_stacked}--\ref{fig:agr_baselines_stacked} show the stacked bars of error patterns. As can be observed, the general trends are the same for both sentence embedding types, which is why we comment only the OpenAI results.
For \textit{spray/load}, the larger number of errors in the train type I/ test type III is dominated by Alt-NP, Alt-PP and LexPrep errors, grammatically correct sentences, which indicate, however, that the sequence has not been learnt. In this difficult setting other lexical (LexPrep) and structural (NoEmb) errors  are also present. The other configurations see a much more pronounced presence of SSM errors, the mistakes due to wrong syntax-semantic mappings.
For CoS and OD, in English and Italian, the dominant error is 
of a semantic nature. \textsc{I-Int} indicates that the intransitive target structure has been identified, but with an incorrect semantic mapping of the subject. For example, for the target sentence derived from the transitive sentence \textit{The student broke the vase}, the chosen candidate is \textit{The student broke} rather than \textit{The vase broke}. In Italian, the range of mistakes is greater, and ranges over almost all semantic kinds, especially when training on type I.  Instead, the use of the morpho-syntactic, reflexive-like element, \textit{si} is essentially learned. These results shows that the syntactic information is learned. 
For agreement, in all languages, the dominant mistakes are WN1 and WN2, mistakes in the grammatical number of the intervening attractors. These are sentences that obey the rule of agreement but show that the rules governing the sequence have not been learnt, like in the \textit{spray/load} case.  OpenAI, whose performance is overall worse,  also shows a small but widely spread presence of agreement errors on the main elements, a true grammar mistake.

\subsection{Discussion}

The results on the baselines show that the task is solvable. It is important to establish the task can be solved with reasonable performance, to be able to say that the analyses of the inner representations we perform in the following section derive from the attempt of the architecture to solve this task. We also show that these baselines are reliable as they carry across sentence embedding types, and so are the profiles of the range of errors. The error distribution shows clearly --especially for \textit{spray/load} and agreement, and across languages--- that the grammatical properties have been learnt, but that the  logic of the sequence in  the puzzle has not.
From a more linguistic interpretation of the errors, the mistakes seem to indicate a tendency to prefer an Agent interpretation for the first argument ---a well-documented bias also in human speakers \cite{huber2024NoL} --- and to apply  local solutions in agreeement.

\section{Identifying sentence structure with BLMs}
\label{sec:sentences}

LLMs work at the level of the token. But language, and the linguistic abstractions and regularities encoded in BLMs, are expressed by more semantically meaningful units, such as morphemes, words or phrases. 
We study whether sentence embeddings contain information about the chunk structure of the corresponding sentences. We compress the embeddings into a lower-dimensional representation in a dedicated VAE-like architecture. If information about the structure is encoded, we could capture it in a lower-dimensional representation that could be used to solve the BLM problems better than with the uncompressed sentence representations.

\paragraph{Creating the data}

From each of the BLM tasks, we extract sentences from the context portion of each instance, and the associated patterns. 
For the patterns of the \textit{spray/load} alternation data, see  Figures \ref{ALT-ATL} and \ref{ATL-ALT}; for the CoS and OD datasets, see  Figure \ref{tab:COStemplateBLM}. 
We randomly select 4004 sentences for each of the verb alternation datasets, uniformly spread over the seven patterns.
From the agreement data (for each language separately), we obtain sentences with structure of the form: {\rm NP (PP$_1$ (PP$_2$)) VP}, where the parentheses surround optional structure. Each chunk can have singular or plural form, with agreement between the first NP (the subject) and the VP. This organisation leads to 14'336 sentences with one of 14 patterns, from which we randomly choose 4004 sentences uniformly distributed over the 14 patterns.

In this way, we obtain four datasets from the agreement data -- in English, French, Italian and Romanian --, two datasets from the \textit{spray/load} data (in English, for each of the two alternations), four datasets from the CoS and OD verbs data (in English and Italian). Each such dataset consists of 4004 instances, uniformly split over the patterns represented in the data. These are split into train:dev:test -- 2576:630:798.
%
We build sentence representation as averaged token embeddings obtained from a pretrained Electra model \cite{clark2020electra}\footnote{\textit{google/electra-base-discriminator}}, reshaped as two-dimensional arrays (32x24) \cite{nastase-merlo-2023}.

\paragraph{Sentence structure through a VAE model}

The latent structure representation of the sentence data is investigated using a variational encoder-decoder, similar to a variational auto-encoder, but the decoder does not reconstruct the input\footnote{Throughout the experiments we do abbreviate this to {\it VAE}.}. The encoder consists of a CNN layer with a 15x15 kernel, which is applied to a 32x24-shaped sentence embedding, followed by a linear layer that compresses the output of the CNN into a latent layer of size 5. The decoder mirrors the encoder, and unpacks a sampled latent vector into a reshaped two-dimensional 32x24-array sentence representation, which indicates that patterns are encoded periodically, and are best detected with a 15x15 kernel.

A training  instance consists of a triple $(in, out^+, Out^-)$, where $in$ is an input sentence with embedding $e_{in}$ and chunk structure $p$, $out^+$ is a sentence with embedding $e_{out^+}$ with  same chunk structure $p$, and $Out^- = \{s_k| k=1,N_{negs}\}$ is a set of $N_{negs}=7$ sentences with embeddings $e_{s_k}$, each with chunk pattern different from $p$ (and different from each other). The input $e_{in}$ is encoded into a latent representation $z_i$, from which we sample a vector $\tilde{z}_i$, which is decoded into the output $\hat{e}_{in}$. 

We enforce that the latent encodes the structure of the input sentence by using a max-margin loss function, shown in (\ref{maxMloss}) and (\ref{maxM}), to push for a higher score with the sentence that has the same chunk pattern as the input ($e_{out^+}$) than the ones that do not ($E^- = \{e_{s_k}| e_{s_k}=embedding(s_k), s_k \in Out^-$\}).
At prediction time, the sentence from the $\{out^+\} \cup Out^-$ options that has the highest score relative to the decoded answer is taken as correct.
\begin{equation}
\label{maxMloss}
    {loss}_{(e_{in})} = maxM(\hat{e}_{in},e_{out^+}, E^-) + KL(z_i||\mathcal{N}(0,1))
\end{equation}
\begin{equation}
\label{maxM}
    maxM(\hat{e}_{in},e_{out^+}, E^-) = max(0,1-score(\hat{e}_{in}, e_{out^+})+ \frac{\sum_{e_{s_k} \in E^-} score(\hat{e}_{in}, e_{s_k})}{N_{negs}})
\end{equation}

\paragraph{Probing experiments}

To assess whether the correct patterns of chunks are detected, (i) we  analyze the output of the system, in terms of average F1 score over three runs; (ii) we  analyze the latent layer through latent traversal, to see how the confusion matrices change depending on what elements of the latent layer are masked and we perform clustering, to determine whether chunk patterns are encoded in the latent vectors; (iii) we  visualize the projections of the input onto the latent layer, to see whether inputs with the same structure are projected into the same neighbourhood, indicating that their their latent representations cluster.

\textit{F1 Scores}
In a binary evaluation,
the system achieves a very high average positive class F1 score (and standard deviation) over three runs, as 
indicated in Table \ref{tab:F1-scores}.  This result means the system has built a sentence representation whose chunk pattern is closest to the chunk pattern of the input.

\begin{table}[h!]
\small
\centering
\setlength{\tabcolsep}{1.8mm}
\footnotesize
\begin{tabular}{l|rr|rr|rr|rrrr}
& \multicolumn{6}{c|}{Verb Alternations}   & \multicolumn{4}{c}{Agreement}  \\
&\multicolumn{2}{c}{Spray-load} & \multicolumn{2}{c}{CoS} & \multicolumn{2}{c|}{OD}  & \multicolumn{4}{c}{ }  \\
&\tiny{(ALT-ATL)}  & \tiny{(ATL-ALT)}  & En              & It                & En      & It              
& English & French  & Italian & Romanian  \\ \hline
F1 & 0.951      & 0.952      & 0.909   &  0.960     &   0.951     &  0.975    
  & 0.959      &  0.951  & 0.917  & 0.896   \\
SD  &  (0.006)     &  (0.005)     &  (0.015)   &   (0.002)    &    (0.033)    &   (0.004)    
  &  (0.010)     &   (0.011) &  (0.007) &  (0.003)  \\
\end{tabular}
    \caption{F1 scores of whether the correct patterns of chunks are detected in the sentences of the  BLM datasets.}
    \label{tab:F1-scores}
\end{table}

\textit{Pattern-level evaluation}
To understand how chunk information is encoded on the latent layer, we perform latent traversals: for each instance in the test data, after encoding it, we modify the value of each unit in the latent layer with ten values in its min-max range, based on the training data, and decode the answer.

\begin{figure}
    \centering
    \includegraphics[width=0.6\textwidth]{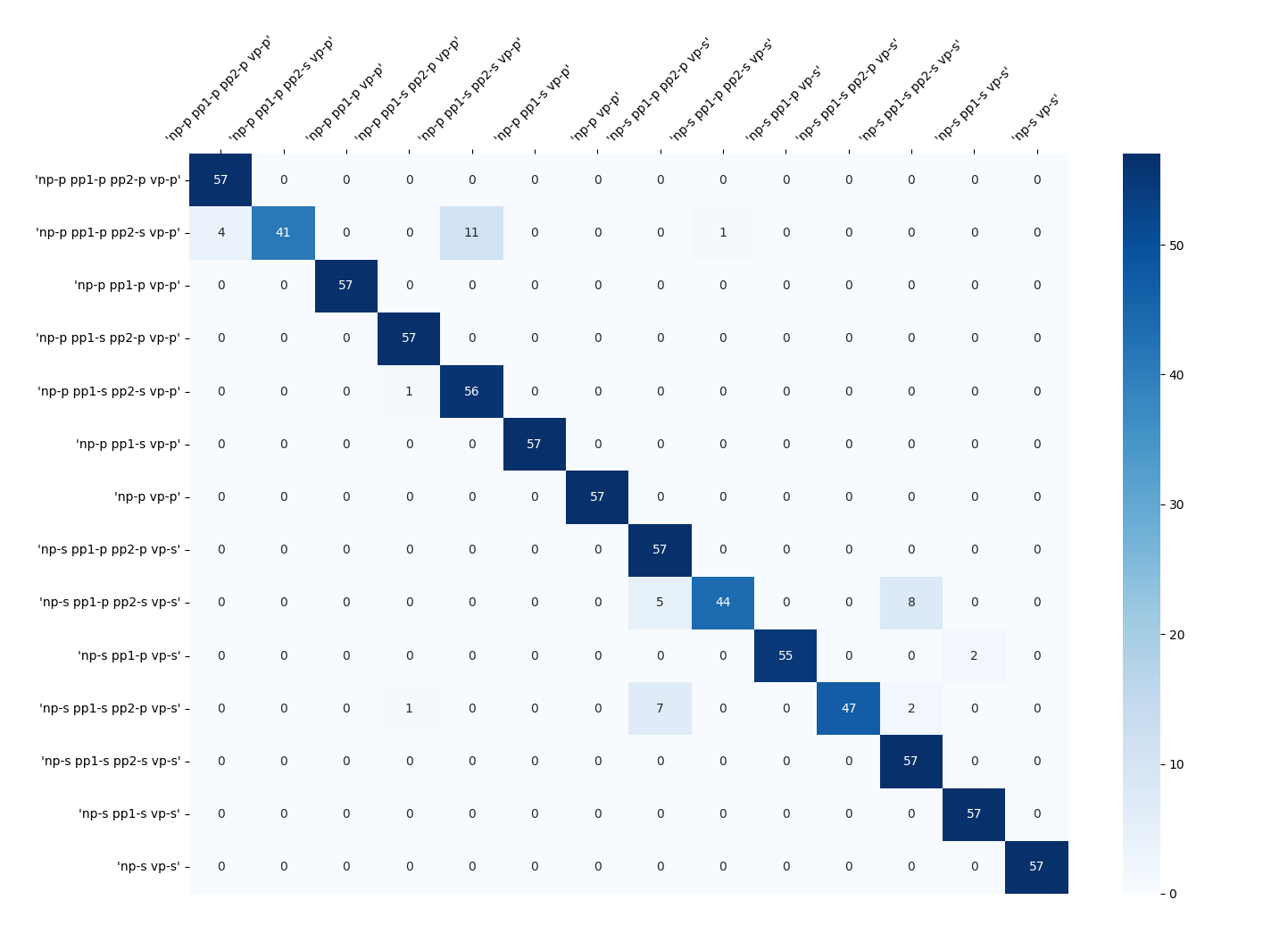}
    
    \vspace{1mm}
    \includegraphics[width=0.8\textwidth]{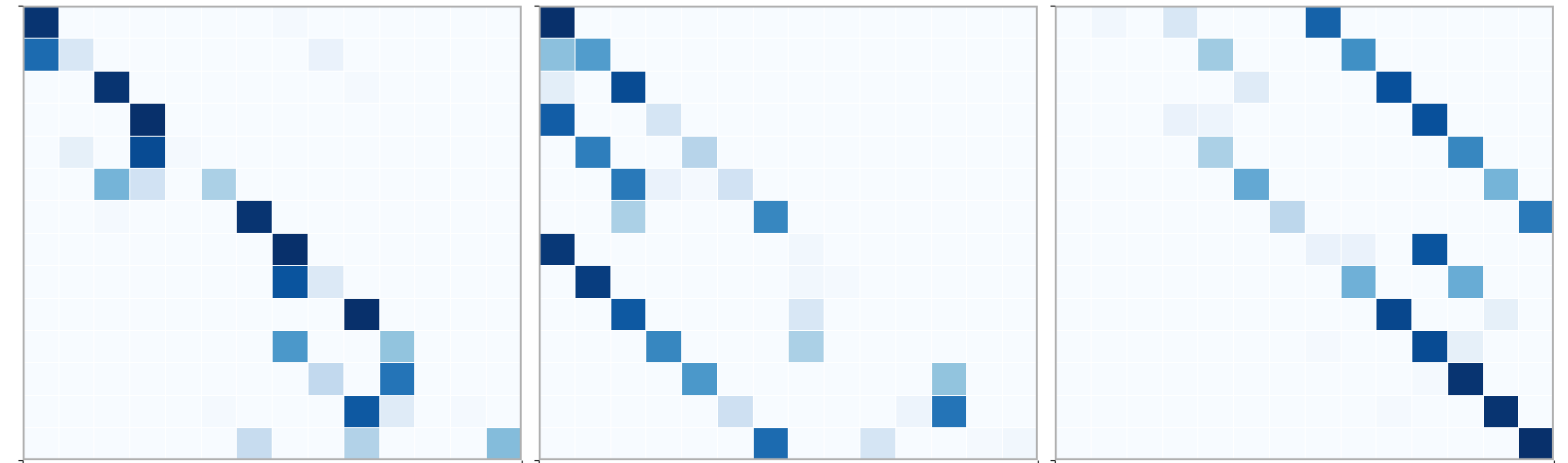}

    \vspace{-3mm}
    \caption{Confusion matrix (top) and 
    sample of effects of latent traversal in terms of pattern-level evaluation (bottom). }
    \label{fig:sentence_latents-agr}
    \vspace{-3mm}
\end{figure}

The pattern-level evaluation for the French agreement data, presented as a confusion matrix based on the pattern information for $out^+, Out^-$ in  Figure \ref{fig:sentence_latents-agr} (top), shows that all patterns are detected with high accuracy. 
The confusion matrices presented as heatmaps in  Figure \ref{fig:sentence_latents-agr} (bottom)  show that specific changes to the latent vectors decrease the differentiation among patterns, as expected if chunk pattern information were encoded in the latent vectors. Changes to latent unit 1 cause patterns that differ in the grammatical number of $pp2$ not to be distinguishable (left matrix). Changes to latent units 2 and 3 lead to the matrices in the middle and right of the figure, where patterns that have different subject-verb grammatical number are indistinguishable.

\begin{figure}
    \centering
\includegraphics[width=\textwidth]{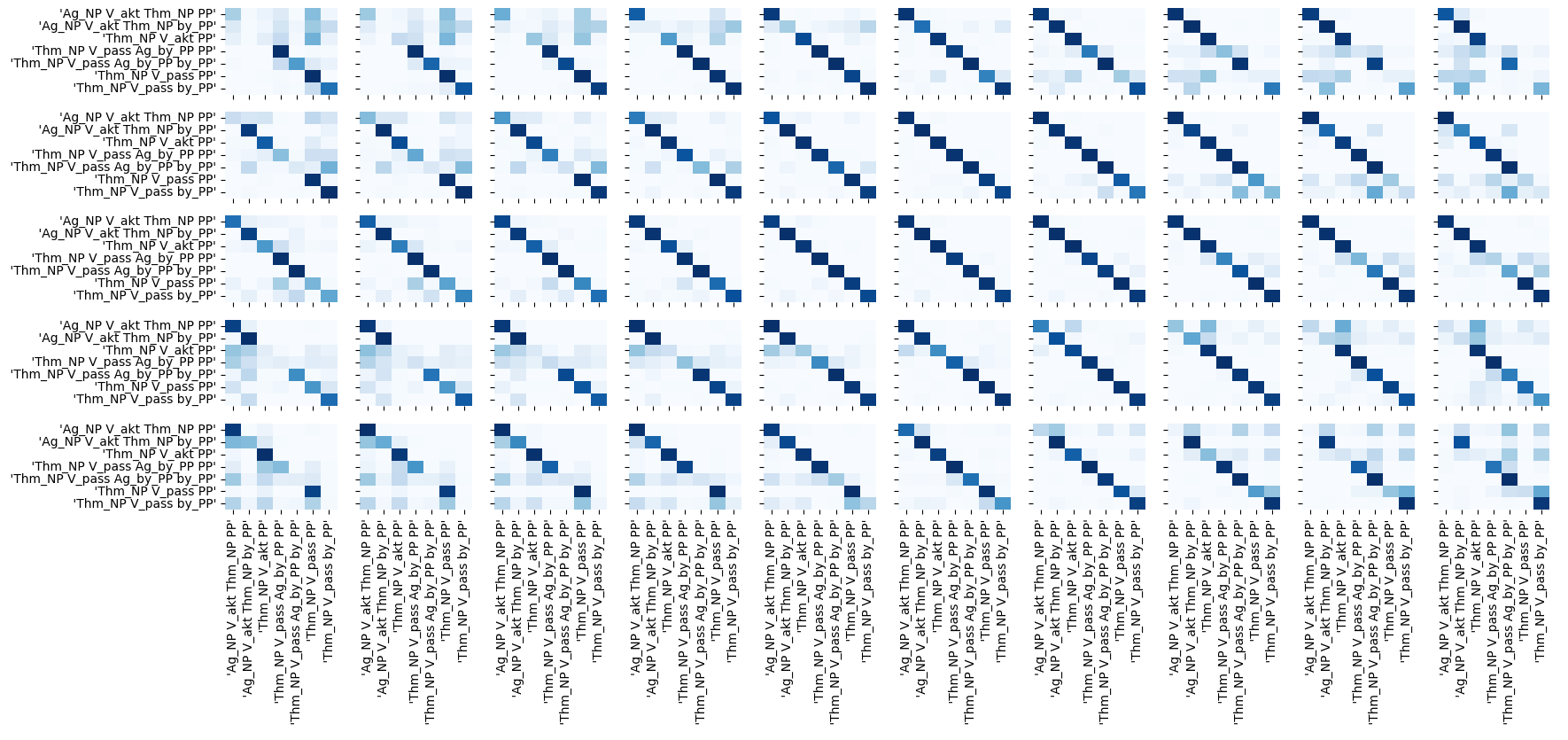}
    \caption{ Latent layer encoding of pattern information:
    sample of effects of latent traversal in terms of pattern-level evaluation (English, Change of state verbs). Each of the five rows of matrices refers to one latent unit and each of the ten columns of matrices refers to one of the value modifications. Each latent unit is traversed separately.}
    \label{fig:sentence_latents-cos}
    \vspace{-3mm}
\end{figure}

The pattern-level evaluation for the English Change-of-State data are shown in Figure \ref{fig:sentence_latents-cos}, each row represents a latent unit and each column a value of the latent unit that is being modified.\footnote{It should be noted that  the figure needs to be read horizontally, as each unit is modified independently of the others, so vertical cross-sections are not meaningful.}
Not all units show informative changes. We concentrate on the first and fifth unit (first and fifth row). The left and right extremes of each row show more variation, so we will comment these matrices. 
The first unit (first row) captures the fact that the first three sentences are in the active form while the others are in the passive form and the notion that the subject is an Agent or a Theme.
The low values of the latent unit (left) confuse Agent subjects with Theme subject (consider for example the top left corner of the first matrix of the first row), while the high values of the unit (right) conversely confuse Themes for Agents (consider for example the bottom left corner of the first matrix of the first row). Confusion between the third pattern and the sixth pattern in these panels also indicates lack of distinction of verb voice (active or passive), a correlated phenomenon to the change of semantic role of the subject typical of passive voice.
The fifth unit (fifth row) captures information about the different PPs. We observe confusions in the left bottom corner of the panel for the low values of the latent unit and in the right top corner of the panel for the high values. This expected symmetry confirms that the unit captures a specific kind of information. Specifically, the information that appears to be lost concerns the distinction between the agentive \textit{by}-PP and the other PPs.
Combining these results we can conclude that the distinction between active and passive has been learnt, with its correlates of 
change from Agent to theme subject and the existence of an agentive PP which has properties differently from a normal PP. These properties are used in the change-of-state BLM to indicate knowledge of the causative/inchoative alternation  (Theme subject but active voice, sequence pattern based on different kinds of PPs).

\begin{figure}
    \centering
    \begin{tabular}{ccc}
    ALT-ATL & EN CoS & Multilingual Agr\\
    \includegraphics[width=0.27\linewidth]{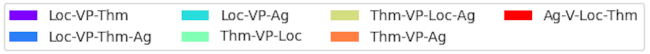}
    &
    \includegraphics[width=0.33\linewidth]{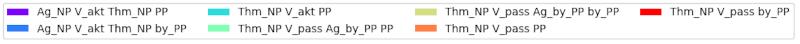}
    &
    \includegraphics[width=0.33\linewidth]{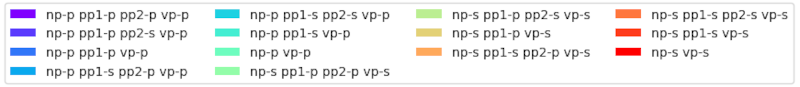}
    \\
    & & {\tiny EN: \ding{108}, FR: \ding{54}, IT: \ding{58}, RO: \ding{72}} 
    \\
    \includegraphics[width=0.27\linewidth]{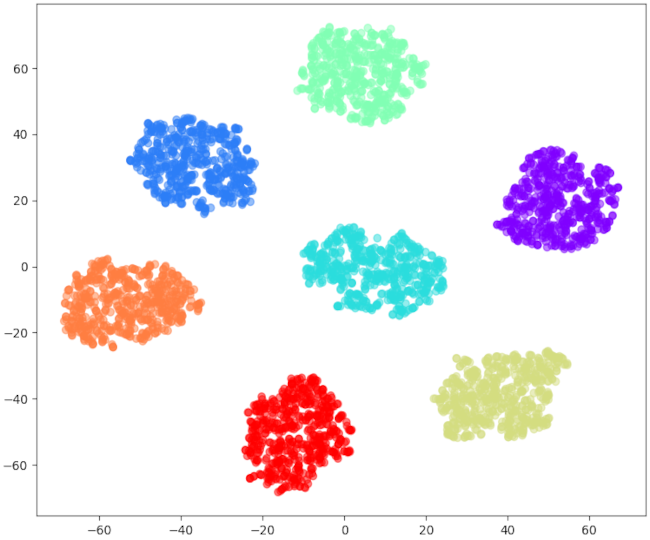}
    &
    \includegraphics[width=0.33\linewidth]{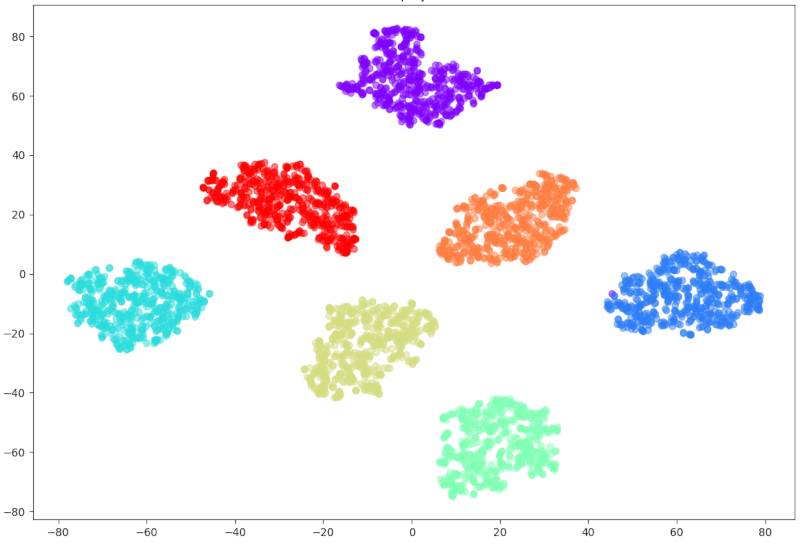}
    &
    \includegraphics[width=0.33\linewidth]{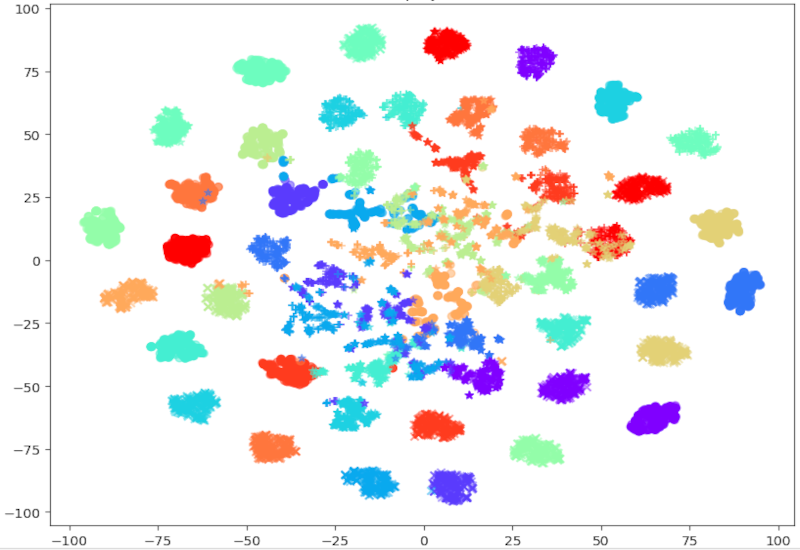}
    \\
    \includegraphics[width=0.27\linewidth]{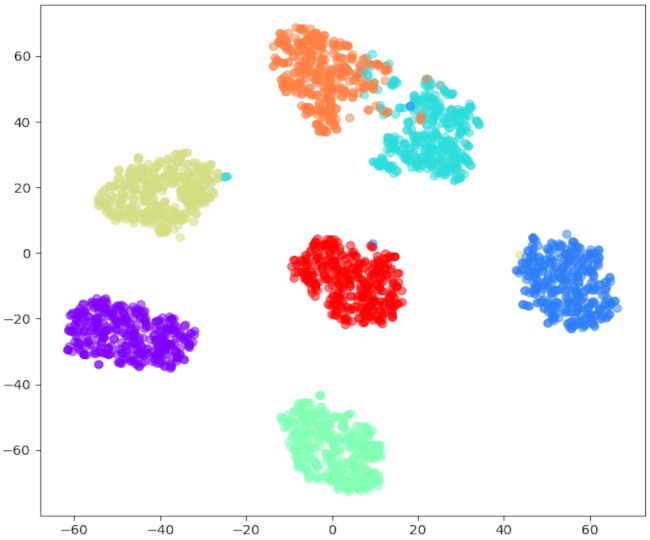}
    &
    \includegraphics[width=0.33\linewidth]{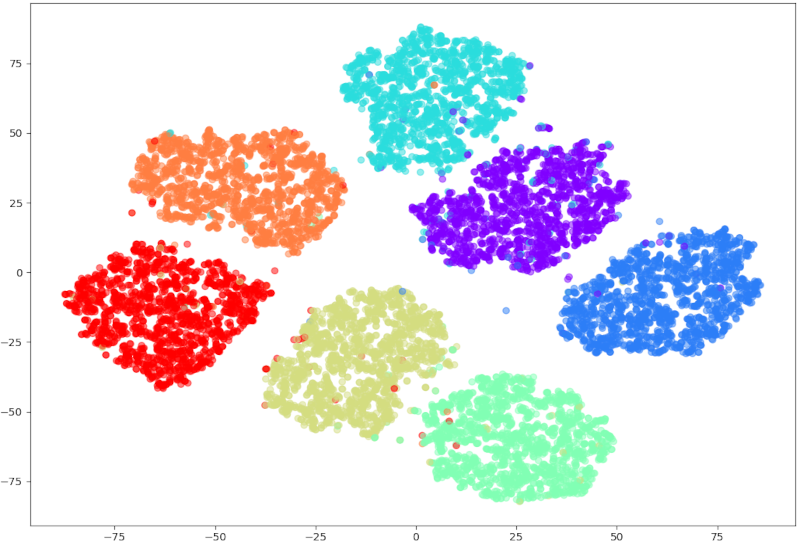}
    &
    \includegraphics[width=0.33\linewidth]{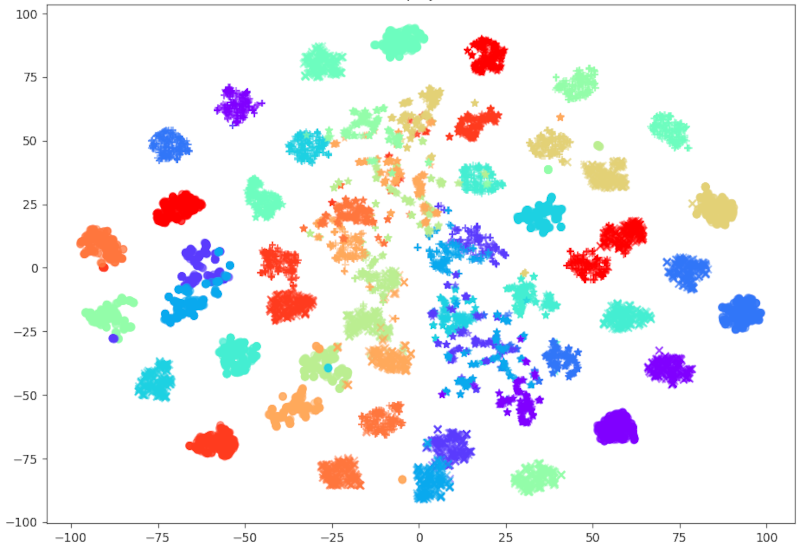}
    \end{tabular}
    \caption{Sentence structure analysis from data of  Type I of \textit{spray/load}, change-of-state (English), and multilingual Agreement (top OpenAI embeddings, bottom Electra embeddings).}
    \label{fig:partsinsentences}
\end{figure}

\textit{Latent Layer}
As additional evidence that chunk information is present in the latent layer, we plot the projection of the latent vectors in two dimensions (Figure \ref{fig:partsinsentences}). The plot shows a very crisp clustering of latents that correspond to input sentences with the same chunk pattern, despite the fact that some patterns differ by only one attribute (e.g., the grammatical number) of one chunk.

\paragraph{Discussion}

This set of experiments shows that information about the chunk patterns that constitute the basic objects and attributes that are relevant for the BLM tasks can be distilled from the sentence representations. Despite receiving input at the level of tokens, LLMs encode cues about higher level structure, which we made more overt through the VAE learning formalism. 
The investigation of this section has explored the  internal structure of single sentence embeddings. The investigations in the next section will study BLMs sequences of sentences, and how they cast light on systematicity.

\section{BLM structure to identify systematicity}
\label{sec:systematicity}

In the general discussion of core basic properties of human language, systematic compositionality occupies a central place \citep{donatelli2023compositionality}. Natural languages compose elements so that  meaning is a compositional function of the meaning of the parts. Moreover and importantly,
they do so systematically, like in  a mathematical system of equations --- a set of equations where common solutions hold. Systematicity, then, is a statement about the relationship we find across sentences.

By its nature and by design,  BLM is an implicit test of systematicity. For example, consider the chunking and shared values of the chunks,  colour-coded in the figures illustrating the BLM templates (Figures \ref{fig:template-matrices}, \ref{ALT-ATL}, \ref{ATL-ALT}, \ref{tab:COStemplateBLM} and \ref{fig:blm-roll-eg}): these chunks and their properties (semantic role mappings and agreement features) carry across sentences in a  systematic way.
To solve a BLM, the chunks and their relevant properties should be detected, and the relations between them across sentences must be  learnt. If they are, then we can conclude that the model learns systematically. 

The experiments in Section \ref{sec:sentences} show that chunks and their properties are identifiable in sentence embeddings. We embed this process into a dedicated BLM puzzle solver, 
to replace the sentence representations with their compressed version that provide chunk information more overtly, and test whether a system can detect 
the internal cross-sentence systematic structure of the BLM sequence. We  compare these specialised VAEs to the  FFNN OpenAI and Electra baselines described in previous sections. We expect that learning will improve, compared to these baselines, because it is precisely the systematic cross-sentence structure that provides the organisational abstraction needed for informative compression and to achieve systematic learning.

\subsection{Two-level VAE}

\begin{figure}[h]
    \centering
    \includegraphics[width=0.7\textwidth]{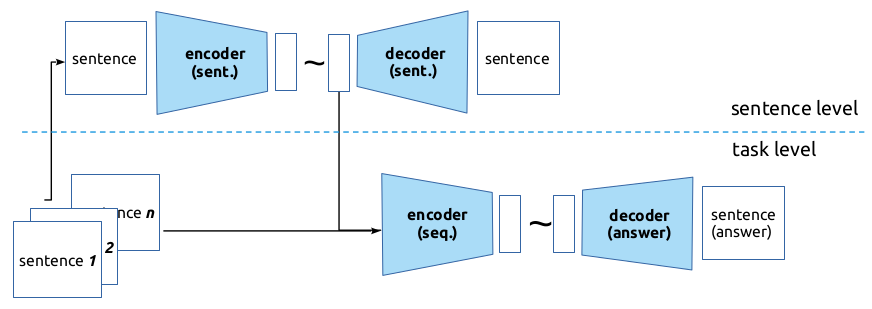}
    \caption{A two-level VAE-based system: the sentence level learns to compress a sentence into a useful representation for solving the BLM problem at the task level.}
    \label{fig:2levelVAE}
\end{figure}

The specialised two-level VAE models the BLM problem directly.
BLM problems encode a linguistic phenomenon in a sequence of sentences that have regular and systematic structure 
\cite{carpenter1990one}. 
We model the process of solving a BLM by using the two-level VAE encoder-decoder architecture illustrated in Figure \ref{fig:2levelVAE}. 
One level of this two-level VAE encoder-decoder  (we call it sentence level) compresses the sentence embeddings of each of the sentences in the input sequence into a small latent vector (size 5), as described in Section \ref{sec:sentences}. The sampled latent representations are then used as the representations of the sentences in the input sequence. The second level (we call it task level) receives  this sequence representation as input and decodes an answer
based on the internal structure of the sentences and their systematic relations expressed in the sequence. 

Recall that an instance for a BLM task consists of a tuple, comprising a sequence of sentences $S = \{s_i |i = 1, 7\}$ as input, and an answer set with one correct answer $a_c$, and several incorrect answers $a_{err}$.  The two-level system only receives the BLM instances as input, which must be processed through the sentence-level part first to obtain the new and compressed sentence representations. 
The sentence-level VAE requires an instance to be a triple $(s_i, out^+_i, Out^-_i)$. For each input to the BLM system, the instances for the sentence-level VAE are generated dynamically, during processing. From the sequence of seven sentences in an input BLM context, we construct seven instances for the sentence-level VAE, one for each sentence: $s_i \in S$ with embedding $e_{s_i}$, $out^+_i = s_i$, and $Out^-_i = \{s_k| s_k \in S, s_k \ne s_i\}$ with embeddings $E^-_i = \{e_{s_k}| k =1, N_{negs}\}$.

The loss at the sentence level is computed as described in equation (\ref{maxMloss}) and repeated here, for ease of reference, in equation (\ref{eq7}).
The task level receives  this sequence representation  as input, it compresses it into a new latent layer, and the sampled vector is then decoded into a sentence representation that  best matches the representation of the correct answer.
 The loss at the task level is computed relative to the answer set $\mathcal{A}$ with the corresponding embeddings set $E_A$, and the correct answer $a_c$, of the task, as in \ref{combinedloss}.
The two levels are learned together and the loss of the system combines the loss signal from the two levels, as in  (\ref{combined-levelsloss}).

\vspace{-0.5cm}
\begin{equation}
    \label{eq7}
{loss}_{sent}(e_{s_i}) = maxM(e_{s_i}, e_{out^+_i}, E^-_i) 
+ KL(z_i|\mathcal{N}(0,1)).
\end{equation}

\vspace{-1cm}
 \begin{equation}
    \label{combinedloss}
 {loss}_{task}(e_S) = maxM(e_S, e_{a_c}, E_A\setminus{e_{a_c}}) + KL_{seq}(z_S|\mathcal{N}(0,1)).
\end{equation}

\vspace{-1cm}
\begin{equation}
    \label{combined-levelsloss}
{loss}_S= \sum_{s_i \in S} {loss}_{sent}(e_{s_i}) + {loss}_{task}(e_S).
\end{equation}

\begin{figure}
    \centering
    \includegraphics[width=0.8\linewidth, trim={0, 0, 0, 1cm}, clip]{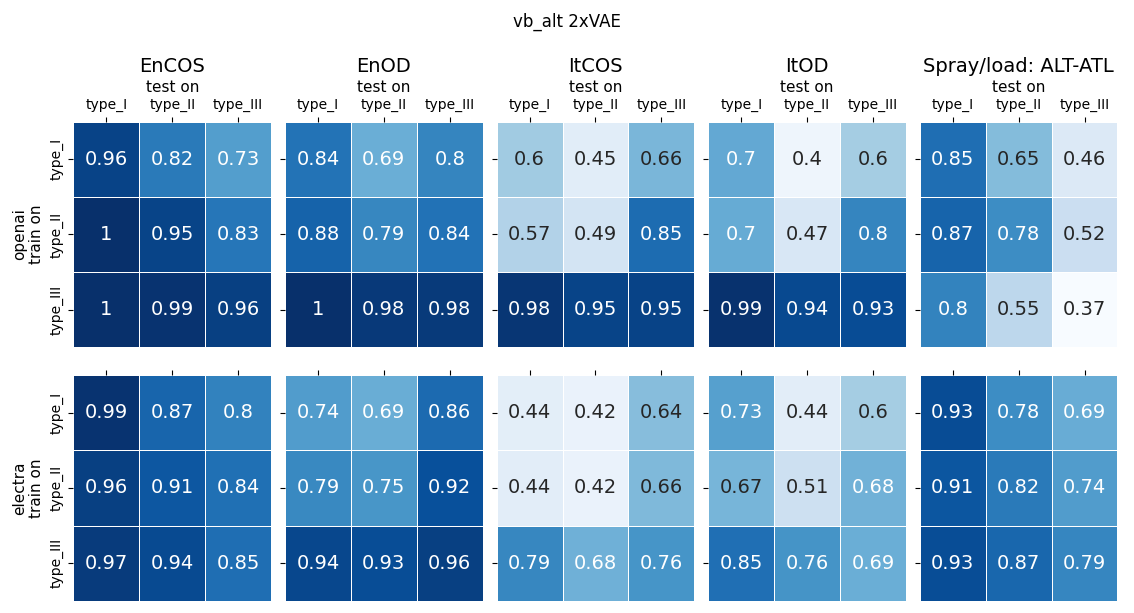}

    \caption{F1 verb alternations (two-level VAE, OpenAI top, Electra bottom): CoS (En,It), OD (En,It), spray/load (En).}
    \label{fig:two-level-VAE-verb_alt}
\end{figure}

\begin{figure}
    \centering
    \includegraphics[width=0.8\linewidth, trim={0, 0, 0, 1cm}, clip]{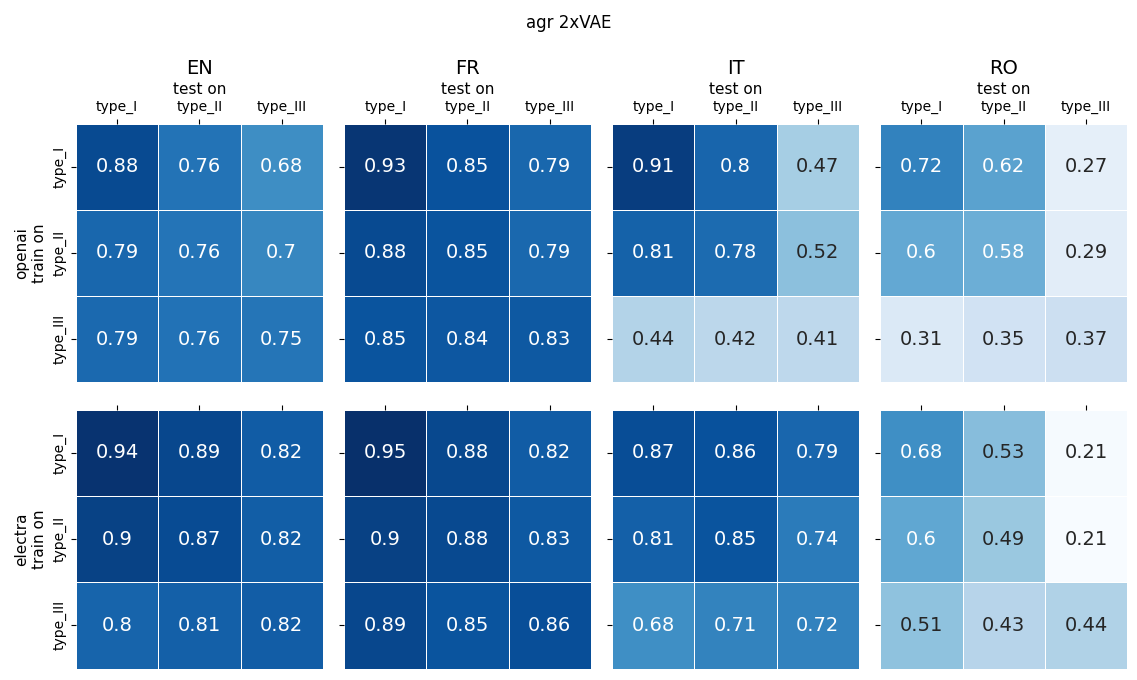}
    
    \caption{F1 Agr (two-level VAE, OpenAI top, Electra bottom) for four languages.}
    \label{fig:agr_f1_two-level-VAE1}
\end{figure}

\begin{figure}
    \centering
    \includegraphics[width=0.7\linewidth, trim={0, 0, 0, 1cm}, clip]{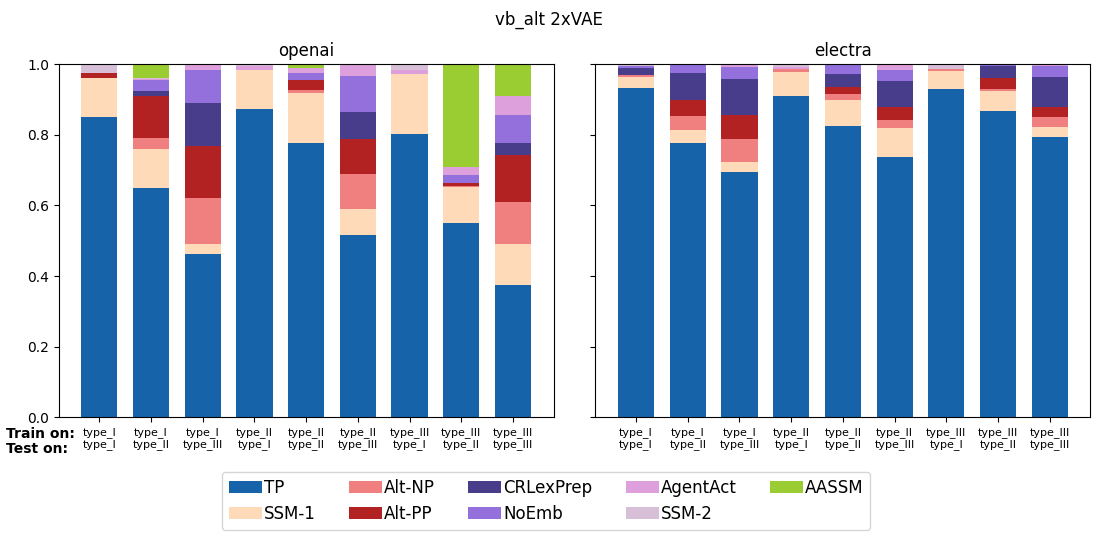}
    \caption{Answer type distribution (correct + errors) for verb alternations (two-level VAE): \textit{spray/load} in English}
    \label{fig:two-level-VAE-verb_alt_stacked}
\end{figure}

\begin{figure}
    \centering
   
    \includegraphics[width=0.5\linewidth, trim={0 0 0 1cm}, clip]{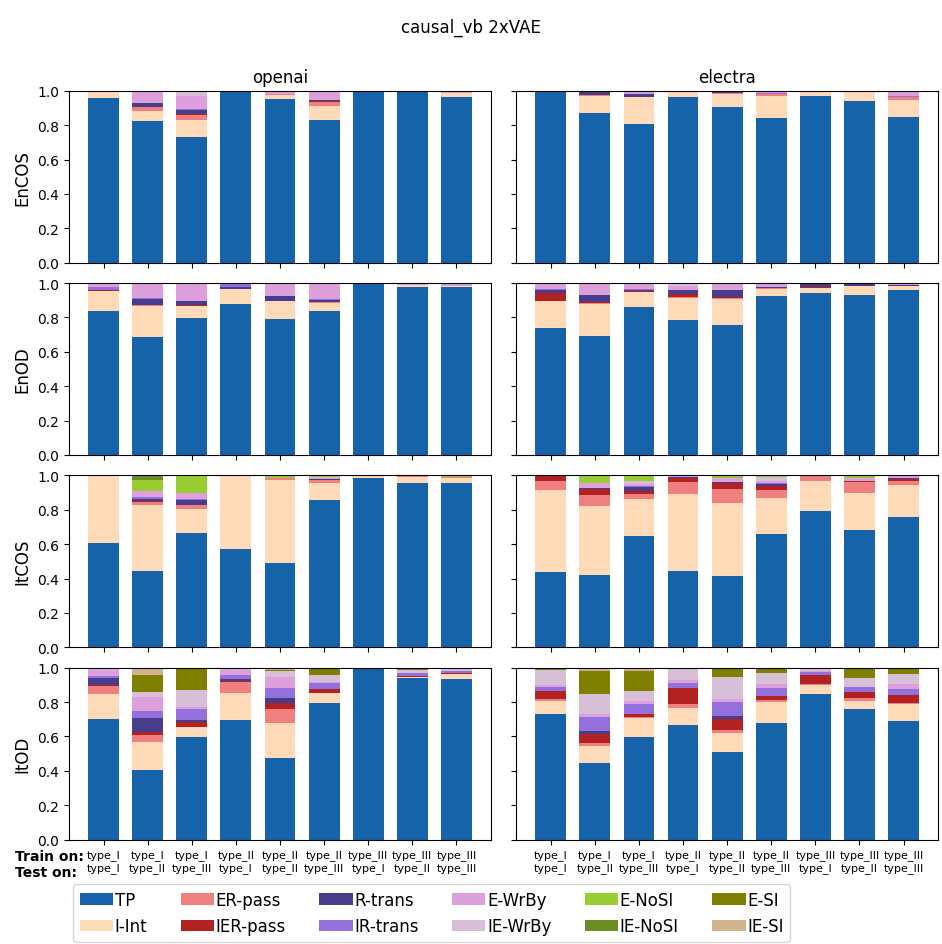}
    \caption{Answer type distribution (correct + errors) for verb alternations (two-level VAE, OpenAI left and Electra right): COS and OD in English and Italian}
    \label{fig:two-level-VAE-cause_vb_stacked}
\end{figure}

\begin{figure}
    \centering
    \includegraphics[width=0.5\linewidth, trim={0, 0, 0, 1cm}, clip]{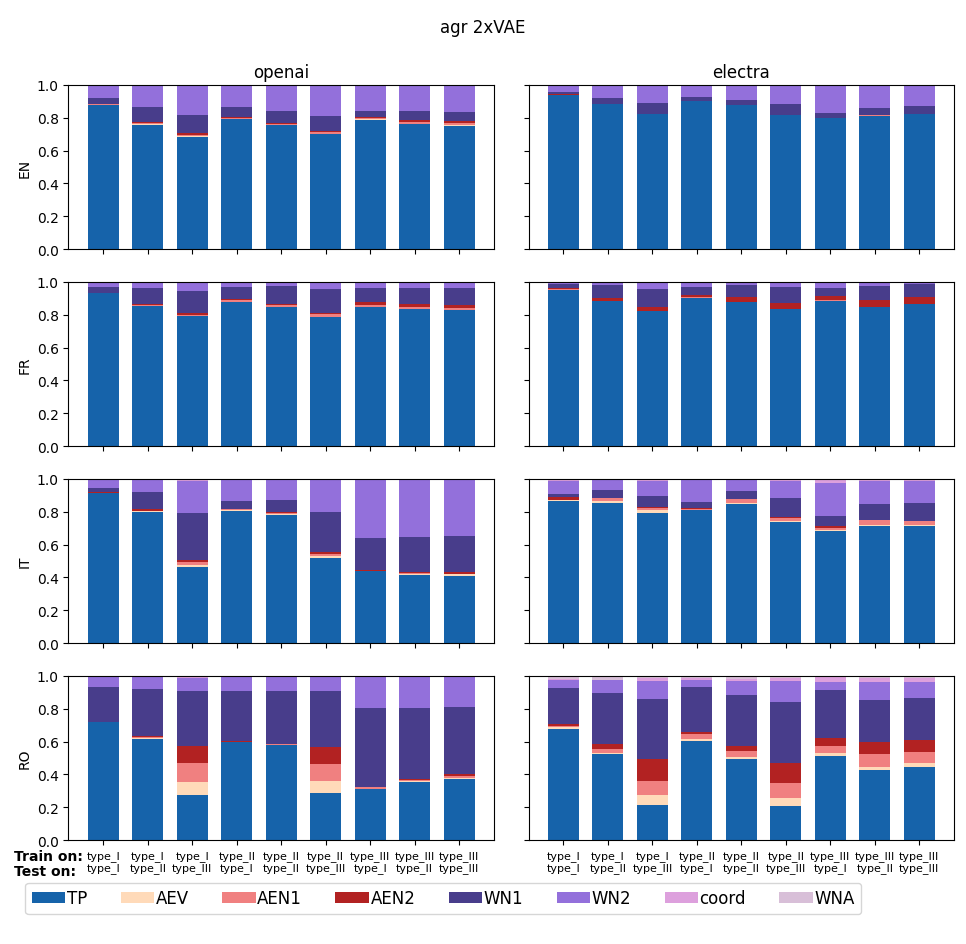}
    \caption{Answer type distribution (correct + errors) for Agr (two-level VAE, OpenAI left, Electra right) for 4 languages}
    \label{fig:agr_f1_two-level-VAE}
\end{figure}

\pagebreak
\subsection{Experiments and discussion}

We run experiments on the BLMs for the \textit{spray/load} alternation and the causative alternation, shown in Figure \ref{fig:two-level-VAE-verb_alt}, and for the BLM on agreement in Figure \ref{fig:agr_f1_two-level-VAE1}, to see if the two-level VAE identifies the range of systematic syntactic and semantic structural properties of these BLMs. Solving the agreement task requires  long-distance structural information, while verb alternations involve both structural information concerning phrases and semantic information, with syntactically similar phrases (those that have the same grammatical function) playing different roles in a sentence. 

The results in terms of average F1 scores and error analyses provide several insights. 
Consider the comparison of results in Figures \ref{fig:two-level-VAE-verb_alt} and \ref{fig:agr_f1_two-level-VAE1} to the baselines in Figures \ref{fig:baseline_resultsValt} and \ref{fig:agrFFNN} in section 6. In Agreement and Object drop BLMs, experiments trained or tested on type III data see an increase in F1 scores. 
In the baseline experiments, we expected that variability in the input data would lead to better generalizing and more robust models. Specifically,  we expected type III data to lead to higher results than type I data. This did not happen. However, using the 2-level VAE system, with a more overt sentence structure, leads to increase results of the models trained on type III data. 
This result indicates that the two-level VAE compression helps recognising the underlying patterns beyond the variability introduced by lexicalisation. A weaker improvement for Type III is also seen in \textit{spray/load}. The already good results of the Change-of-state alternation see a noticeable error rate reduction thanks to the two-level architecture. 

Compare also the error distributions shown in Figures \ref{fig:two-level-VAE-verb_alt_stacked}, \ref{fig:two-level-VAE-cause_vb_stacked} and \ref{fig:agr_f1_two-level-VAE} to the respective baselines (Figures  \ref{fig:vb_alt_baselines_stacked},     \ref{fig:cause_vb_baselines_stacked} and     \ref{fig:agr_baselines_stacked}).
For \textit{spray/load}, we see that most errors are concentrated in the condition where type I is used in training and type III in testing, with a fairly even distribution of mistakes across four  main error types  (Alt-NP/PP, Lexprep, noEmbed). For the CoS and OD English verbs, we see that, like for the baseline, the dominant mistake is I-Int, a sequence error. The agreement data has particularly strong sequence patterns. Comparing the agreement error distributions in Figures \ref{fig:agr_f1_two-level-VAE} and  \ref{fig:agr_baselines_stacked}, we note a large decrease in grammatical errors (AEV, AEN1, AEN2). This result indicates that the automatically-induced sentence structure contributes to solving the task. 

The performance of the two-layer system confirms that relevant structure, linguistic objects and their properties, and the systematic regularities among them are correctly detected and used to solve the tasks. We see, instead, mostly extra-grammatical sequence mistakes.

\section{General discussion and related work}

In this paper, we have introduced Blackbird Language Matrices (BLMs) ---linguistic puzzles developed in analogy to the visual tests called Raven Progressive Matrices (RPMs) \cite{merlo2023}--- and studied the systematic generalisation abilities in neural networks in solving these puzzles. 

We have shown that  BLM datasets are richly structured and support many different types of investigations, at both the sentence and matrix levels. The context-answers setup supports counterfactual investigations of possible types of errors: language errors, reasoning errors, and their interactions \cite{an-etal-2023-blm,nastase-merlo2023,samo-etal-2023-blm}. 
Specifically, in the investigation of object induction, we show that we can detect linguistically justified objects, like phrases and constituents, in sentence embeddings from pretrained language models. Furthermore, their task-relevant properties -- like semantic roles in verb alternations, or number in agreement -- are also detected. Our two-step BLM task-solver shows how we can detect symbolic structures in the distributed sentence representations and use them for solving linguistic puzzles \cite{boleda-2025-llms}.
The regular syntactic forms and the systematic semantic properties support investigations on systematicity in neural networks. 
BLMs exists for several tasks and different languages, enabling multi-tasks and multi-language comparative studies \cite{nastase-etal-2024-multitask,nastase-etal-2024-multilingual}.
Finally, each BLM problem is a linguistic paradigm and can be seen as a tool for linguistic investigation of specific phenomena.

Unlike other attempts to create textual versions of RPMs, BLMs are natural stimuli and not simple direct transcriptions of visual stimuli  \cite{webb2023emergent} (see below for a critique), 
nor are  they auxiliary abstractions of stimuli in the visual domain \cite{hu-etal-2023-context}. Instead, BLMs are matrices developed specifically to learn language-related problems and study deeper formal and semantic properties of language, through a process of linguistic paradigm understanding.  While they encode true linguistic problems, BLMs are structured like RPMs: they encode properties of objects and their attributes, and to be fully solved one has to discover the underlying rules that tie all the components in a logical sequence.  Also, like 'progressive' matrices, they are organised into increasing levels of difficulty.

As such, our proposal contributes to a rich debate on what data and tests  should be developed to advance and measure machine intelligence. We also contribute to a deeper analysis of  LLMs' linguistic knowledge and we provide alternative proposals for structured datasets, natural and synthetic, to study generalisation and systematicity.

\paragraph{\textbf{Measuring intelligence}} In a forward-looking paper on machine intelligence, \citet{chollet2019} highlights the shortcoming of benchmarking. Benchmarks, while very useful,  trigger task-oriented solutions, while reaching truly flexible and general solutions requires developing systems that can handle \textit{unknown unknowns}, systems that can adapt to novel tasks (thus unseen problems), novel to the machine, but also the developer.
In Chollet's view, the concrete proposal that should take up the challenge is the ARC dataset. It is based on the tradition of psychometric tests, like RPMs, operationalising the notion of intelligence in a measurable way, to allow for comparisons among individuals, be they humans or machines. ARC puzzles are an RPM-like block world (whose test set has never been released).
ARC problems constitute the most similar dataset in intent, realization, and measuring abilities to the BLM dataset, but they differ because ARC is a visual dataset, it is a simplified artificial grid world and, as such, unlike BLMs, does not correspond to any explicit natural class of problems.

A recent survey and position paper \citep{MAHOWALD2024517}, weighing in on the more cognitive side and in the comparison to humans,  has attempted to reconcile some of the apparently contradictory results on LLMs by positing a loose distinction supported by  neurological evidence: LLMs'  formal knowledge of language is good, but their functional knowledge is not. That is, LLMs know the rules of grammar well, but do not understand the rules that govern the use of language. 
ARC and BLM can contribute to this debate. It has been claimed that ARC is a test of analogical ability and fluid intelligence \cite{mitchell2023abstraction}. Results on BLMs seem to indicate that LLMs solve BLM problems, like \citet{carpenter-ea1990} subjects, analytically and not analogically \cite{merlo2023}. They are able to identify objects and attributes that describe the formal properties of language, but have a hard time with the logical structure of the template,  the reasoning and use of language part. Explicit manipulation of analogical structure of the paradigm does not seem to be useful \cite{jiang2025analogicalstructureminimalcontextual}. 
Like ARC, BLMs also belong to the trend of data and task creation that define operational measures of intelligence. As advocated for ARC, BLMs also feature a considerable component of original manual structure building. This manual novelty sets BLMs apart from other datasets meant to measure linguistic skills, which are collections of pre-existing handmade examples (Holmes) \cite{waldis-etal-2024-holmes}, or are large collections of minimal pairs that differ by only one or two words, a task that does not require problem-solving ability (BLiMP) \citep{warstadt-etal-2020-blimp-benchmark}.

\paragraph{\textbf{Contrastive datasets}}

Contrastive datasets, such as the BLiMP benchmark,  are widely used in testing models' linguistic knowledge. 
BLiMP is a collection  of binary contrastive  minimal pairs across several linguistic phenomena, gleaned from the linguistic literature. It aims to be a benchmark of large-scale grammatical abilities. It spans 67 paradigms, grouped in twelve larger phenomena. It mainly covers surface morphological phenomena (agreement), and basic phenomena core to the generative tradition (argument structure, island constraints, binding, control, NPI and quantifiers). 
Holmes  is another extensive derivative dataset collection ---i.e., a collection of preexisting datasets----- created by aggregating more than 200 previously existing individual datasets, also including BLiMP. The datasets are grouped thematically into macro-categories of  syntax, morphology, semantics, reasoning, and discourse
phenomena. This contrastive approach has been extended to many other languages for some limited set of phenomena (MultiBLiMP) \cite{jumelet2025multiblimp10massivelymultilingual}.
Contrastive data, including minimal and contrastive pairs, are important in evaluating and understanding the capabilities and limitations of NLP models \citep{mccoy-etal-2019-right,gardner-etal-2020-evaluating, ide2025make} and the sources of their results \citep{mccoy-etal-2019-right}. 

Other research in the literature, however, has suggested that complex linguistic phenomena require more elaborate structures than mere pairs for accurate model assessment \cite{vamvas-sennrich-2021-limits}.
BLMs extend the logic of minimal pairs a step further into more complex matrix-like structures of contrastive examples that  vary systematically along multiple dimensions, both in the context and answer set and  allow for systematic testing across multiple linguistic dimensions simultaneously. Other contrastive datasets are often used with simple, binary classifier probes,
a process that neither provides insights into how the LLM has encoded linguistic information, nor whether it uses any such information for its decisions. 
Instead, a BLM answer set allows the researcher to investigate \textit{why} certain answers are chosen, by looking at the mistakes and setting up multiple contrasts (Jiang and Merlo, in progress).

\paragraph{\textbf{Synthetic datasets}}

In the world of progressive matrices as human intelligence tests (\citet{raven1938} and following), the stimuli  are designed by hand, but the original disentanglement research in vision  using RPMs to train neural networks deliberately employed some structured generative model to create a large number of questions \citep{wang2015ijcai,barrett2018pmlr,zhang2019cvpr,zhu2006sig}. 
More generally, interest in synthetic data generation has been growing since the advent of transformer architectures and large language models, driven by the need to augment training datasets with new data points that did not exist in the original datasets, particularly  with low-resource languages and varied tasks (information extraction, question-answering, reasoning, semantic similarity prediction) \citep{wu-etal-2019-conditional,kumar-etal-2020-syntax,panikolaou-etal-2020-dare,yang-etal-2020-generative, mohapatra-etal-2021-simulated-chats,schick-schutze-2021-generating, huang-etal-2025-targa}.
More recent data-synthetic methods rely on prompt-based techniques on LLMs for zero-shot and few-shot learning tasks. This line of work is popular because it can generate coherent and contextually adequate synthetic data without the need for extensive model retraining \citep{bonifacio-etal-2022-inpars,dai-etal-2022-promptagator,ye-etal-2022-zerogen,meng-etal-2022-generating, huang-etal-2025-targa}. 
Techniques like the zero-shot Chain of Thought (CoT) prompting \cite{wei2022chain,kojima2022large} are used to distil complex reasoning capabilities from LLMs \cite{shao-etal-2023-synthetic,shum-etal-2023-automatic, diao-etal-2024-active}, reverse task prompting can produce high-quality synthetic data that is both relevant and contextually varied \cite{josifoski-etal-2023-exploiting}.

It has however been argued that adding more
training data is a way of artificially ``buying'' specialised performance without inducing any generalization power \citep{chollet2019}.
For this reason, the BLM problems illustrated above exhibit a mixture of natural data, manual linguistic intervention, and synthetic techniques to augment data sizes and  variability. The seed data  often come from natural sources, such as corpus data. The templates are built by hand. The augmentation process is semi-automated, with several steps of manual validation.  While this interleaving of hand-construction and semi-automated processes is labour-intensive and expert-dependent, it also guarantees the variability, naturalness that is necessary to create data  and obtain results related to generalisation abilities,  and that other algorithmically-generated synthetic datasets cannot guarantee.

\paragraph{\textbf{Datasets for generalisation}}

Because of their mixture of systematically manipulated structure, and semi-manual creation, the BLM datasets are well-suited to investigate generalisation \cite{Hupkes2023,vansteenkiste2020disentangled,zheng-lapata-2022-disentangled}.

Our dataset strives for similar goals to other highly controlled structured datasets, such as the COGS dataset \cite{kim-linzen-2020-cogs}, which aims at providing out-of-distribution test cases, in structure and lexical variation.
The COGS dataset, however,  determines the combinatorics by design  that the network needs to find, imposing preexisting hard independence assumptions generated by a CFG in the test set. 
Our BLM approach is more in the spirit of hidden representation learning. We do not require that the network has explicitly learnt new generative ways of combining elements. But we encourage representations that learn soft constraints, in the form of disentangled representations that correspond to the generative underlying factors and lexical generalisation (through the different types of lexical variability in the matrix) \cite{merlo-etal-2023-blackbird}.

Because of their more naturalistic nature, BLMs are apt to probe inner representations.  The datasets we propose in this paper have the necessary properties: they focus on specific linguistic phenomena, they display lexical and structural variation, and include known confounding factors for the targeted phenomena. They are, then, close to the work that investigates network representations \cite{lasri-etal-2022-probing, kann-etal-2019-verb,yi-etal-2022-probing}.

\paragraph{\textbf{Computer vision and multi-modal}}

The BLM datasets provided here are related to those used in the literature on disentanglement in computer vision \cite{ACMSurveyRPMComputerVision2025}. For example,  \citet{vansteenkiste2020disentangled} developed a dataset for computer vision similar to RPMs. They evaluate the usefulness of the representations learned for abstract reasoning and note that learning disentangled representations leads to faster few-shot learning. 
RAVEN is another widely used dataset for computer vision that encodes very simplified RPMs  \cite{chi2019RAVEN}, where each attribute is independent of the other as it is generated by a context-free grammar. 
\cite{webb2023emergent}
report on multi-modal results on analogical tests with LLMs that are based on RPMs. One of the tests is the Digit Matrices problem set (see, for example, their figure on page 1528), which they define as `a novel text-based matrix reasoning task designed to emulate Raven’s SPM problems'. 
With a process very similar to the one used in \citet{webb2023emergent}, \citet{hu-etal-2023-context}
find, in fact,  that LLM are not very apt at analogical reasoning in the visual domain, based on the RAVEN dataset. To help, they encode perception features (shape, colour, position) in language and solve the problem as a multi-modal task with great improvement.  

These papers differ from BLMs in that they either introduce techniques that do not map directly to language data or textual data that change the nature of the RPM problem. On the one hand, generating data by a context-free grammar, as in RAVEN,  is  not appropriate in naturalistic language data, where sentences are subject to overarching grammatical and lexical constraints.
On the other hand, in \citet{webb2023emergent},   the task 
uses text as a modality, but has none of  the properties of text or language, it is a textual digital translation of the visual Raven matrices. The textual matrices are an expedient to be able to provide the matrix to the  LLM. In \citet{hu-etal-2023-context}, the  fact that language 
abstraction provides a big improvement shows that the encoding in language solves a part of the problem. Unlike what is claimed, in both these works, it is unlikely that the improvement is due to the better abstraction of the perceptive features, but rather to the explicit encoding of the abstract dimensions of the problem, due to naming, and the consequent identification of objects and attributes, thus solving essential intermediate problems.

BLMs, instead, provide the task that truly corresponds to a language version of RPMs, and provides the input as raw sentences. The model needs to identify the correct chunks in the input sentences, detect the relevant attributes and, based on objects and attributes, construct a logical sequence that brings to the solution.

\section{Conclusions} 

This article describes the Blackbird Language Matrix task (BLM), the BLM datasets, their benchmarking and more detailed experiments. It supports the point of view that curated, structured datasets
give rise to complex, multi-dimensional problems that are more naturalistic, and support multi-faceted investigations of properties of language and large language models.
Blackbird Language Matrices are such curated datasets. We have shown they can be solved at a good, but challenging level of performance, as a linguistic puzzle in their own right, for several linguistic rules, in more than one language, with simple baseline models or, at better performance levels, with more tailored models. 
These curated, but naturalistic datasets are useful to answer some core questions about current large language models abilities. We have, for example, shown that their representations contain  the grammatical objects and attributes relevant to solve a linguistic task. We have also shown these solutions are reached by detecting systematic patterns across sentences.   
Because they present a curated, articulated structure, because they comprise both learning contexts and expected answers, and because they are partly built by hand, BLMs fall in the category of datasets that can support investigations of general intelligence, and be useful in investigating more deeply \textit{why} large language models ---these aliens the current AI debate now anthropomorphises into almost human-like entities--- behave the way they do.

\section*{Acknowledgments}

We gratefully acknowledge the support of this work by the Swiss National Science Foundation, through grant SNF Advanced Grant  TMAG-1\_209426 to PM.

\appendixsection{Appendices}
\subsection{Mixture BLM Template and example}

\label{app:template}

\begin{minipage}{0.5\textwidth}
\footnotesize
\begin{tabular}{llll} \hline
\multicolumn{4}{c}{\textsc{Context}}\\\hline
1 & \textcolor{blue}{Sg}\textcolor{violet}{M}-\textcolor{brown}{CPas}   & & \textcolor{blue}{Sg}\textcolor{violet}{M}-\textcolor{brown}{CnPr}\\
2 & \textcolor{red}{Pl}\textcolor{violet}{M}-\textcolor{brown}{CPas}   & & \textcolor{red}{Pl}\textcolor{violet}{M}-\textcolor{brown}{CnPr}\\
3 & \textcolor{blue}{Sg}\textcolor{teal}{F}-\textcolor{purple}{CTrPas}  & & \textcolor{blue}{Sg}\textcolor{teal}{F}-\textcolor{purple}{CnPas}\\
4 & \textcolor{red}{Pl}\textcolor{teal}{F}-\textcolor{purple}{CTrPas}  & & \textcolor{red}{Pl}\textcolor{teal}{F}-\textcolor{purple}{CnPas}\\
5 & \textcolor{blue}{Sg}\textcolor{purple}{M}-\textcolor{orange}{IFut}   & attractor\textcolor{red}{PL}\textcolor{teal}{F} & \textcolor{blue}{Sg}\textcolor{purple}{M}-\textcolor{orange}{IFut}\\
6 & \textcolor{red}{Pl}\textcolor{purple}{M}-\textcolor{orange}{IFutAnt} & attractor\textcolor{blue}{Sg}\textcolor{violet}{M} & \textcolor{red}{Pl}\textcolor{purple}{M}-\textcolor{orange}{IFut} \\
7 & \textcolor{blue}{Sg}\textcolor{teal}{F}-\textcolor{orange}{IFut}    & attractor\textcolor{red}{Pl}\textcolor{violet}{M} & \textcolor{blue}{Sg}\textcolor{teal}{F}-\textcolor{orange}{IFut}\\
8 & ??\\\hline
\multicolumn{4}{c}{\textsc{Answers}}\\\hline
A & \textcolor{red}{Pl}\textcolor{teal}{F}-\textcolor{orange}{IFutAnt} & attractor\textcolor{blue}{Sg}\textcolor{teal}{F} & \textcolor{red}{Pl}\textcolor{teal}{F}-\textcolor{orange}{IFut}\\
B & \textcolor{blue}{Sg}\textcolor{teal}{F}-\textcolor{orange}{IFut}    &               & \textcolor{blue}{Sg}\textcolor{teal}{F}-\textcolor{orange}{IFut}\\
C & \textcolor{blue}{Sg}\textcolor{violet}{M}-\textcolor{purple}{CTrPas}  &               &  \textcolor{blue}{Sg}\textcolor{violet}{M}-\textcolor{purple}{CnPas}\\
D & \textcolor{red}{Pl}\textcolor{violet}{M}-\textcolor{orange}{IFutAnt} &               &  \textcolor{red}{Pl}\textcolor{violet}{M}-\textcolor{orange}{IFut}\\
E & \textcolor{blue}{Sg}\textcolor{teal}{F}-\textcolor{purple}{CTrPas}  & attractor\textcolor{blue}{Sg}\textcolor{teal}{F} & \textcolor{red}{Pl}\textcolor{violet}{M}-\textcolor{purple}{CTrpas}\\
F & \textcolor{red}{Pl}\textcolor{teal}{F}-\textcolor{orange}{IFutAnt} &  & \textcolor{red}{Pl}\textcolor{teal}{F}-\textcolor{orange}{IFut}\\
\hline
 \end{tabular}
\end{minipage}
\begin{minipage}{0.5\textwidth}
\small

\underline{Contexts and Answers patterns}  

(CPas:past subjunctive; CTrPas:perfect past subjunctive; CnPr:present conditional; CnPas:past conditional; IFut:future indicative; IFutAnt:future in the past indicative).

\underline{Properties of correct answer} 

(A): Agr:yes, Number:Pl, Gender:F, attractor:yes, TensePremise:IFutAnt, TenseConsequent:IFut.

\underline{Violations in answer set}: 

(B) - attractor:no, Agr:Pl; 
(C) - Number:Sg, Gender:M, tense:Cpas, attractor: no;
(D) - Gender:M, attractor: no;
(E) - Agr: no, Number:Sg, tense:CtrPas;
(F) - attractor: no.
\end{minipage}

\footnotesize
\begin{tabbing}
\\
\\
\textsc{Context}\\
1 \ \= Se fossi ricco, allora sarei felice.\\
\> {\footnotesize \textit{If I were rich, then I would be happy.}}\\

2 \ \= Se fossero buoni, allora non sarebbero ricchi.\\
\> {\footnotesize \textit{If they were good, then they would not be rich.}}\\

3 \ \= Se fossi stata ricca, allora non saresti stata felice.\\
\> {\footnotesize \textit{If you had been rich.F.SG, then you would not have been happy.F.SG.}}\\

4 \ \= Se foste state buone, allora sareste state ricche.\\
\> {\footnotesize \textit{If you had been good.F.PL, then you would have been rich.F.PL.}}\\

5 \ \= Se sarai bravo, indipendentemente dalle circostanze, allora sarai ricco.\\
\> {\footnotesize \textit{If you are good.M.SG, regardless of the circumstances, then you will be rich.}}\\

6 \ \= Se sarete stati bravi, nonostante tutto, allora sarete ricchi.\\
\> {\footnotesize \textit{If you will have been good.M.PL, despite everything, then you will be rich.}}\\
7 \ \= Se sarai buona, tenendo conto di tanti altri fattori, allora sarai ricca.\\
\> {\footnotesize \textit{ If you are good.F.SG, taking many other factors into account, then you will be rich.}}\\
\end{tabbing}

\begin{tabbing} \textsc{Answers}\\
A \ \= Se sarete state buone, e con un po' di fortuna, allora sarete ricche.\\
\> {\footnotesize \textit{  If you will have been good.F.PL, and with a bit of luck, then you will be rich.F.PL.}}\\
B \ \= Se sarai buona, allora sarai felice.\\
\> {\footnotesize \textit{ If you are good.F.SG, then you will be happy.}}\\
C \ \= Se fossi stato buono, allora fossi felice.\\
\> {\footnotesize \textit{If you had been good.M.SG, then you were.SUBJ happy.}}\\
D \ \= Se sarete stati buoni, allora sarete ricchi.\\
\> {\footnotesize \textit{ If you will have been good.M.PL, then you will be rich.}}\\
E \ \= Se fossi stata buona, tenendo conto di tanti altri fattori, allora fossero stati felici.\\
\> {\footnotesize \textit{ If you had been good.F.SG, taking many other factors into account, then they would have been happy.}}\\
F \ \= Se sarete state buone, allora sarete ricche.\\
\> {\footnotesize \textit{If you will have been good.F.PL, then you will be rich.F.PL.}}\\
\end{tabbing}

\pagebreak
\subsection{Guidelines for validation}
\label{app:guidelines}

\begin{figure}[h!]
    \centering
    \includegraphics[width=0.5\linewidth]{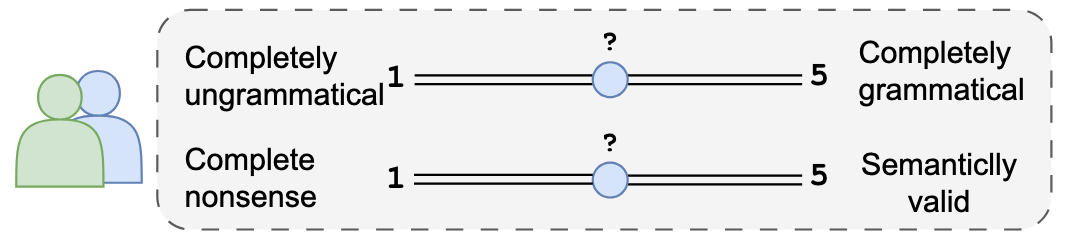}
    \caption{Likert-scale human validation}
    \label{fig:likert-validation}
\end{figure}

\begin{figure}
\small
\paragraph{Introduction}

This validation guideline is designed for participants who are invited to assess the plausibility of simple English sentences. The main goal is to evaluate the quality of the sentences in terms of their \textbf{syntactic} and \textbf{semantic plausibility}. Participants are required to read this guideline and complete the validation process accordingly.

\paragraph{Data Description}

The dataset consists of 192 (out of 1920) sentences with similar syntactic structures, generated using a data augmentation method. These sentences will serve as ``seeds'' for further data creation.

\paragraph{Task Description}

Participants are asked to judge the plausibility of each sentence in the given dataset. The plausibility of a sentence should be evaluated on a scale of 1 to 5, where 1 represents ``strongly not acceptable'' and 5 represents ``strongly acceptable''.

\paragraph{Evaluation Criteria}
\label{app:sec:eval}
Plausibility can be subjective and challenging to define. In this study, we focus on two main aspects:

\begin{enumerate}
  \item \textbf{Overall Syntactic Plausibility}: Assess whether the sentence is grammatically correct or not. Rate the sentence on a scale from 1 to 5, where 1 indicates ``completely ungrammatical'' and 5 signifies ``completely grammatical''. For example, consider the following sentences:
    \begin{itemize}
        \item ``The cat played the piano.'': This sentence is grammatically correct, so it would receive a high score, such as 5 or 5.
        \item ``The cat the piano played.'': This sentence is ungrammatical, so it would receive a low score, such as 1 or 1.
    \end{itemize}
  \item \textbf{Overall Semantic Plausibility}: Evaluate whether the sentence makes sense or not. Assign a score on a scale from 1 to 5, where 1 represents ``complete nonsense'' and 5 denotes ``semantically valid''. For example, consider the following sentences:
    \begin{itemize}
        \item ``The musician played the piano.'': This sentence makes sense, so it would receive a high score, such as 5 or 5.
        \item ``The cat played the piano.'': This sentence is unlikely but not impossible, so it might receive a moderate score, such as 3 or 4.
    \end{itemize}
\end{enumerate}

\paragraph{Requirements}

Participants should have a strong understanding of the English language and be familiar with the principles of syntax and semantics. Ideally, they should be native English speakers to provide accurate and reliable evaluations.

\paragraph{Ethic Concerns}

The data presented in this survey has been generated using large language models (LLMs). LLMs are trained on extensive text data, which may unintentionally incorporate biases present in the training corpus. Consequently, there is a possibility that the generated content could reinforce stereotypes, discriminate against specific groups, or perpetuate societal biases. It is important to note that these data are used strictly for research purposes and do not represent the views or opinions of the researchers conducting this study.

 \caption{Guidelines for  validation}
    \label{fig:guidelines-validation}
\end{figure}

\pagebreak

\pagebreak
\bibliographystyle{compling}
\bibliography{references_cleaned}

\end{document}